\documentclass[10pt,twocolumn]{article}
\usepackage[table]{xcolor}
\pdfoutput=1
\usepackage{cvpr}
\usepackage{times}
\usepackage{graphicx}
\usepackage{amsmath}
\usepackage{amssymb}
\usepackage{array}
\usepackage{enumitem}
\usepackage{multirow}
\usepackage{hhline}
\usepackage{pbox}
\usepackage[numbers,sort]{natbib}
\usepackage{ textcomp }
\usepackage{ appendix }
\usepackage[pagebackref=true,breaklinks=true,letterpaper=true,colorlinks,bookmarks=false]{hyperref}
\usepackage{cleveref}

\cvprfinalcopy

\newcommand{\gd}{\cellcolor{gray!50}}
\newcommand{\gl}{\cellcolor{gray!15}}

\newcommand{\bx}{\mathbf{x}}
\newcommand{\bp}{\mathbf{p}}

\newcommand{\multicell}[1]{\begin{tabular}{@{}l@{}}#1\end{tabular}}
\newcommand{\multicellC}[1]{\begin{tabular}{@{}c@{}}#1\end{tabular}}

\renewcommand{\paragraph}[1]{\par\vspace{0.2em}\noindent{\bf #1}}

\title{Synthetic Data for Text Localisation in Natural Images}
\author{Ankush Gupta\hspace{2em}
    Andrea Vedaldi \hspace{2em}
    Andrew Zisserman\\
    Dept.\ of Engineering Science, University of Oxford\\
    {\tt\small \{ankush,vedaldi,az\}@robots.ox.ac.uk}
}

\begin{document}

\maketitle
\begin{abstract}
In this paper we introduce a new method for text detection in natural
images. The method comprises two contributions: First, a fast and
scalable engine to generate synthetic images of text in clutter.  This
engine overlays synthetic text to existing background images in a
natural way, accounting for the local 3D scene geometry. Second, we
use the synthetic images to train a Fully-Convolutional
Regression Network (FCRN) which efficiently performs text detection
and bounding-box regression at all locations and multiple scales in an
image. We discuss the relation of FCRN to the recently-introduced YOLO
detector, as well as other end-to-end object detection systems based
on deep learning.  The resulting detection network significantly
out performs current methods for text detection in natural images,
achieving an F-measure of 84.2\% on the standard ICDAR 2013 benchmark. 
Furthermore, it can process 15 images per second on a GPU.
\end{abstract}

\section{Introduction}\label{s:intro}

Text spotting, namely the ability to read text in natural scenes, is a
highly-desirable feature in anthropocentric applications of computer
vision. State-of-the-art systems such as~\cite{Jaderberg15c} achieved
their high text spotting performance by combining two simple but
powerful insights. The first is that complex recognition pipelines
that recognise text by explicitly combining recognition and
detection of {\em individual}
characters can be replaced by very powerful classifiers that directly
map an image patch to 
words~\cite{Goodfellow14,Jaderberg15c}. The second is that these
powerful classifiers can be learned by generating
the required training data synthetically
\cite{Wang12,Jaderberg14c}.

\begin{figure}
\begin{center}
   \includegraphics[width=.8\columnwidth]{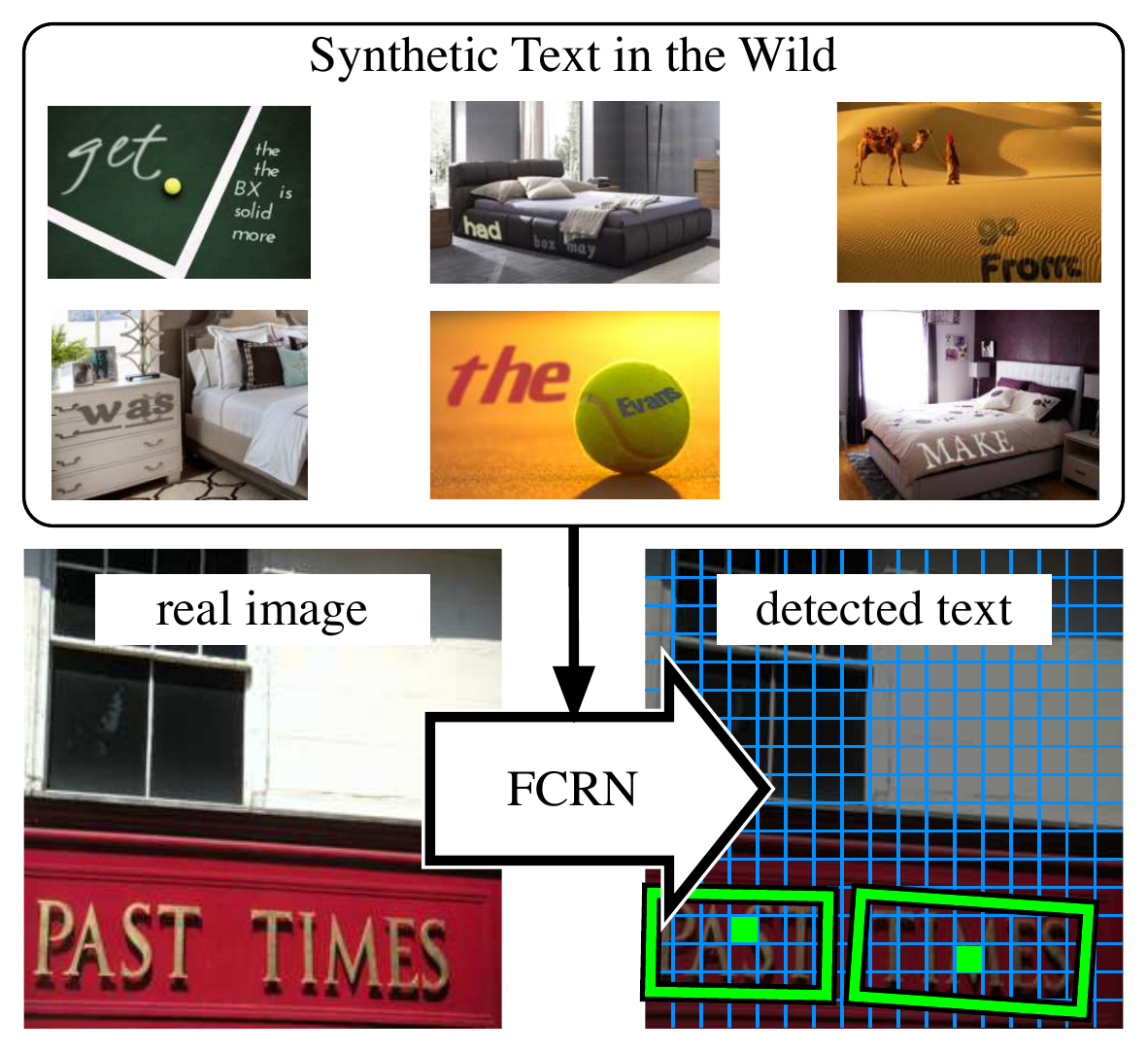}	
\end{center}
\vspace{-1em}
\caption{We propose a Fully-Convolutional Regression Network (FCRN) for
high-performance text recognition in natural scenes (bottom) which
detects text up to 45$\times$ faster than the current
state-of-the-art text detectors and with better accuracy. FCRN is
trained without any manual annotation using a new dataset of synthetic
text in the wild. The latter is obtained by automatically adding text
to natural scenes in a manner compatible with the scene geometry
(top).}\label{fig:teaser}
\end{figure}

While~\cite{Jaderberg15c} successfully addressed the problem of
recognising text \emph{given an image patch containing a word}, the
process of obtaining these patches remains suboptimal. The pipeline
combines general purpose features such as HoG \cite{Dalal05},
EdgeBoxes \cite{Zitnick14} and Aggregate Channel Features
\cite{Dollar14} and brings in text specific (CNN) features only in
the later stages, where patches are finally recognised as specific
words. This state of affair is highly undesirable for two
reasons. First, the performance of the detection pipeline becomes the new
bottleneck of text spotting: in~\cite{Jaderberg15c} recognition accuracy
for correctly cropped words is 98\% whereas the end-to-end text
spotting F-score is only 69\% mainly due to incorrect and missed word region
proposals. Second, the pipeline is slow and inelegant.

\begin{figure}[t]
\centering
\includegraphics[width=\columnwidth]{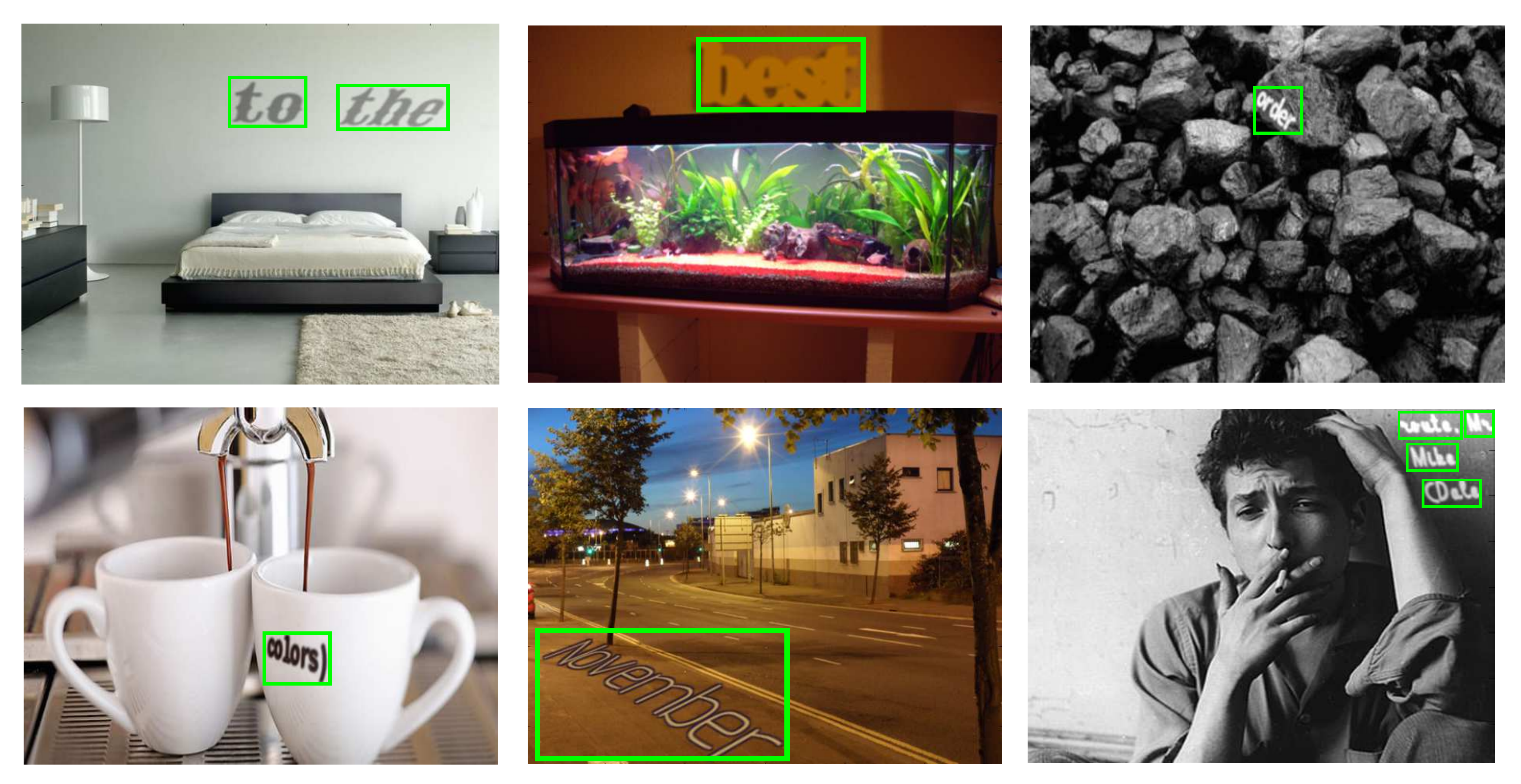}
\caption{Sample images from our synthetically generated scene-text dataset. Ground-truth word-level axis-aligned bounding boxes are shown.}
\label{fig:samplar}
\end{figure}

In this paper we propose improvements similar to~\cite{Jaderberg15c} to
the complementary problem of \emph{text detection}. We make two key
contributions. First, we propose a new method for generating synthetic images of text that \emph{naturally
blends text in existing natural scenes}, using off-the-shelf
deep learning and segmentation techniques to align text to the
geometry of a background image and respect scene boundaries.
We use this method to automatically generate
a new {\bf synthetic dataset of text
in cluttered conditions} (figure~\ref{fig:teaser} (top) and
\cref{sec:synth}). This dataset, called \emph{SynthText in the Wild}~(figure~\ref{fig:samplar}), is suitable for
training high-performance scene text detectors.
The key difference with existing synthetic text datasets such as the
one of~\cite{Jaderberg15c} is that these
only contains word-level image regions and are unsuitable for training
detectors.

The second contribution is a {\bf text detection deep architecture} which is both accurate
 and efficient (figure~\ref{fig:teaser} (bottom) and \cref{sec:bbReg}). We call this
a \emph{fully-convolutional regression network}. Similar to models
such as the Fully-Convolutional Networks (FCN) for image segmentation,
it performs prediction densely, at every image location. However,
differently from FCN, the prediction is not just a class label
(text/not text), but the parameters of a bounding box enclosing
the word centred at that location. The latter idea is borrowed from the
You Look Only Once (YOLO) technique of Redmon~\etal~\cite{Redmon16},
but with convolutional regressors with a significant boost to performance.

The new data and detector achieve
state-of-the-art text detection performance on standard benchmark
datasets (\cref{sec:eval}) while being an order of magnitude faster
than traditional text detectors at test time (up to 15 images per
second on a GPU). 
We also demonstrate the importance of
verisimilitude in the dataset by showing
that if the detector is trained on images
with words inserted synthetically that do {\em not} take account of the scene
layout, then the detection performance is substantially inferior.
Finally, 
due to the more accurate
detection step, end-to-end word recognition is also improved once the
new detector is swapped in for existing ones in state-of-the-art
pipelines. Our findings are summarised in~\cref{sec:conclude}.

\subsection{Related Work}\label{sec:work}

\begin{table}[t]
  \centering
  \begin{tabular}{|c||c|c|c|c|}
    \hline
  \cellcolor{gray!50}& \multicolumn{2}{c|}{\textbf{\cellcolor{gray!50}\# Images}} & \multicolumn{2}{c|}{\textbf{\cellcolor{gray!50}\# Words}} \\   
   \cellcolor{gray!50}  \multirow{-2}{*}{ \textbf{Dataset}} &\cellcolor{gray!15}Train & \cellcolor{gray!15}Test &\cellcolor{gray!15}Train & \cellcolor{gray!15}Test\\
    \hline
    \cellcolor{gray!15}ICDAR \{11,13,15\} & 229 & 255 & 849 & 1095\\
    \hline
    \cellcolor{gray!15}SVT & 100 & 249 & 257& 647\\
    \hline
  \end{tabular}
    \vspace{0.2cm}
    \caption{Size of publicly available text localisation datasets --- ICDAR~\cite{Shahab11, Karatzas13, Karatzas15}, the Street View Text (SVT) dataset~\cite{Wang10b}. Word numbers for the entry ``ICDAR\{11,13,15\}'' are from the ICDAR15 Robust Reading Competition's Focused Scene Text Localisation dataset. }
 \label{table:TextDset}
\end{table}

\begin{figure*}[t]
\centering
\includegraphics[width=7in]{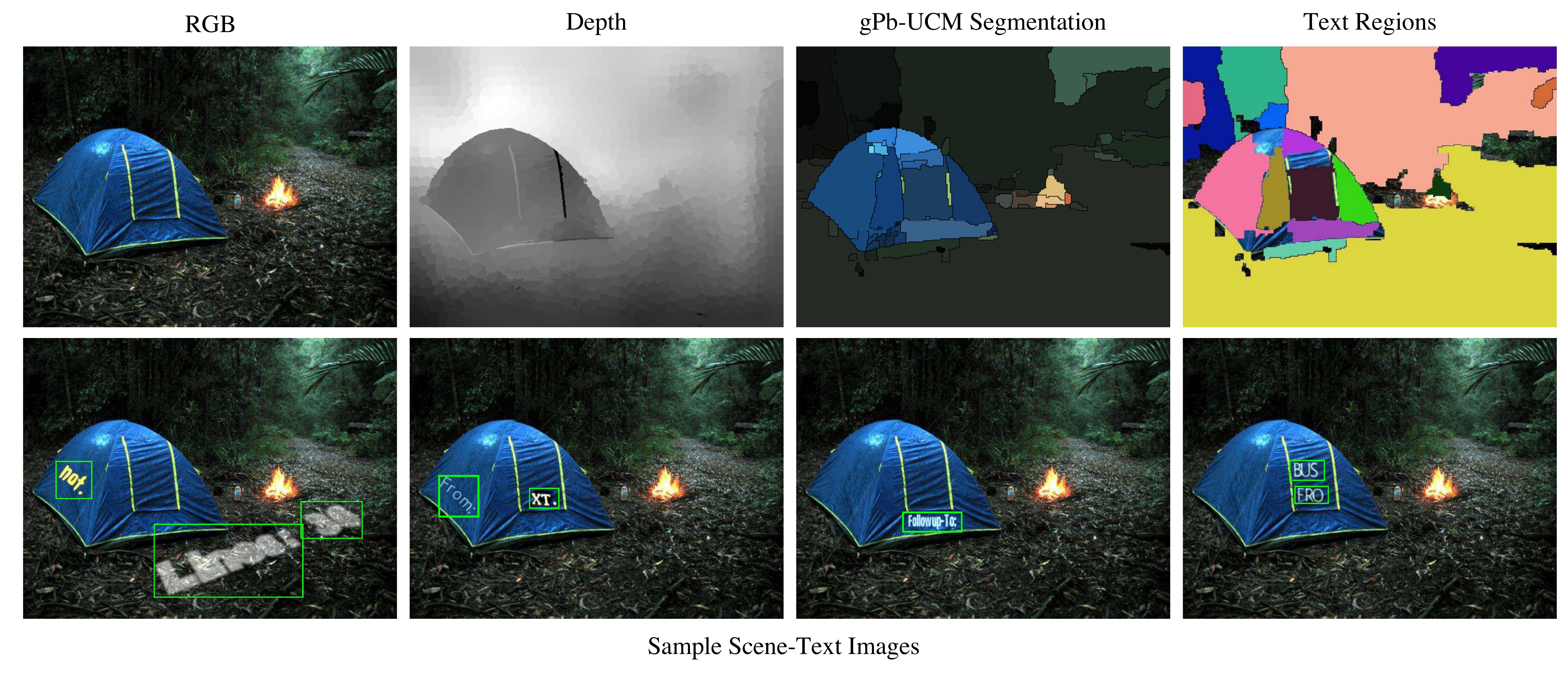}
\caption{(Top, left to right): (1)~RGB input image with no text instance. (2)~Predicted dense depth map (darker regions are closer). (3)~Colour and texture gPb-UCM segments. (4)~Filtered regions: regions suitable for text are coloured randomly; those unsuitable retain their original image pixels. (Bottom): Four synthetic scene-text images with axis-aligned bounding-box annotations at the word level.}
\label{fig:pipeline}
\end{figure*}

\paragraph{Object Detection with CNNs.}  
Our text detection network draws primarily on Long \etal's
Fully-Convolutional network~\cite{Long15} and Redmon \etal's YOLO
image-grid based bounding-box regression network~\cite{Redmon16}.
YOLO is part of a broad line of work on using CNN features
for object category detection dating back to Girshick \etal's
Region-CNN (R-CNN) framework~\cite{Girshick14} combination of
region proposals and CNN features.
The R-CNN framework has three broad stages --- (1) generating
object proposals, (2) extracting CNN feature maps for each proposal,
and (3) filtering the proposals through class specific SVMs.
Jaderberg \etal's text spotting method also uses a similar pipeline for
detection~\cite{Jaderberg15c}. Extracting feature maps for each region
\emph{independently} was identified as the bottleneck by Girshick~\etal
in Fast R-CNN~\cite{Girshick15}. They obtain 100$\times$
speed-up over R-CNN by computing the CNN features once and
pooling them locally for each proposal; they also streamline the
last two stages of R-CNN into a single multi-task learning problem.
This work exposed the region-proposal stage as the new bottleneck.
Lenc \etal~\cite{Lenc15a} drop the region proposal stage
altogether and use a constant set of regions learnt through
K-means clustering on the PASCAL VOC data.
Ren~\etal~\cite{Ren16} also start from a fixed set of proposal,
but refined them prior to detection by using a Region
Proposal Network which shares weights with the
later detection network and streamlines the multi-stage R-CNN framework.

\paragraph{Synthetic Data.} Synthetic datasets provide detailed
ground-truth annotations, and are cheap and scalable alternatives
to annotating images manually. They have been widely used to
learn large CNN models --- Wang \etal~\cite{Wang12} 
and Jaderberg~\etal~\cite{Jaderberg14c} use
synthetic text images to train word-image recognition networks; 
Dosovitskiy \etal~\cite{Dosovitskiy15} use floating chair
renderings to train dense optical flow regression networks.
Detailed synthetic data has also been used to learn generative
models --- Dosovitskiy~\etal~\cite{Dosovitskiy16} 
train inverted CNN models to render
images of chairs, while Yildirim~\etal~\cite{Yildirim15} use
deep CNN features trained on synthetic face renderings to regress
pose parameters from face images.

\paragraph{Augmenting Single Images.} There is a large body of work
on inserting objects photo-realistically, and inferring 3D structure
from single images --- Karsch \etal~\cite{Karsch11} develop an
impressive semi-automatic method to render objects with correct
lighting and perspective; they infer the actual size of objects based
on the technique of Criminisi \etal~\cite{Criminisi99}.
Hoiem \etal~\cite{Hoiem05} categorise image regions into
ground-plane, vertical plane or sky from a single image and use it to
generate ``pop-ups'' by decomposing the image into
planes~\cite{Hoiem05a}.  Similarly, we too decompose a single
image into local planar regions, but use instead the 
dense depth prediction of Liu \etal~\cite{Liu15}.

\section{Synthetic Text in the Wild}\label{sec:synth}

Supervised training of  large models such as deep CNNs, which contain millions of parameters, requires a very significant amount of labelled training data~\cite{Krizhevsky12}, which is expensive to obtain manually.  Furthermore, as summarised in Table~\ref{table:TextDset}, publicly available text spotting or detection datasets are quite small. Such datasets are not only insufficient to train large CNN models, but also inadequate to represent the space of possible text variations in natural scenes --- fonts, colours, sizes, positions. Hence, in this section we develop a synthetic text-scene image generation engine for building a large annotated dataset for text localisation.

Our synthetic engine (1) produces \textbf{realistic} scene-text images so that the trained models can generalise to real (non-synthetic) images, (2) is fully \textbf{automated} and, is (3) \textbf{fast}, which enables
the generation of large quantities of data without supervision.
The text generation pipeline can be summarised as follows (see also Figure~\ref{fig:pipeline}). After acquiring suitable text and image samples (\cref{sec:corpus}), the image is segmented into contiguous regions based on local colour and texture cues~\cite{Arbelaez11}, and a dense pixel-wise depth map is obtained using the CNN of~\cite{Liu15} (\cref{sec:seg}). Then, for each contiguous region a local surface normal is estimated. 

Next, a colour for text and, optionally, for its outline is chosen based on the region's colour (\cref{sec:color}). Finally, a text sample is rendered using a randomly selected font and transformed according to the local surface orientation; the text is blended into the scene using Poisson image editing~\cite{Perez03}.
Our engine takes about half a second to generate a new scene-text image. 

This method is used to generate 800,000 scene-text images, each with multiple
instances of words rendered in different styles as seen in Figure~\ref{fig:samplar}. 
The dataset is available at: http://www.robots.ox.ac.uk/\texttildelow vgg/data/scenetext

\subsection{Text and Image Sources}\label{sec:corpus}
The synthetic text generation process starts by sampling some text and a background image. The text is extracted from the Newsgroup20 dataset \cite{Lang99} in three ways --- words, lines (up to 3 lines) and paragraphs (up to 7 lines). Words are defined as tokens separated by whitespace characters, lines are delimited by the newline character. This is a rich dataset, with a natural distribution of English text interspersed with symbols, punctuation marks, nouns and numbers.

To favour variety, 8,000 background images are extracted from Google Image Search through queries related to different objects/scenes and indoor/outdoor and natural/artificial locales. 
To guarantee that all text occurrences are fully annotated, these images \emph{must not contain text of their own} (a limitation of the Street View Text~\cite{Wang10b} is that annotations are not exhaustive).
Hence, keywords which would recall a large amount of text in the images (\eg ``street-sign'', ``menu'' etc.) are avoided; images containing text are discarded through manual inspection.

\subsection{Segmentation and Geometry Estimation} \label{sec:seg}

\begin{figure}[t]
\centering
\includegraphics[width=\columnwidth]{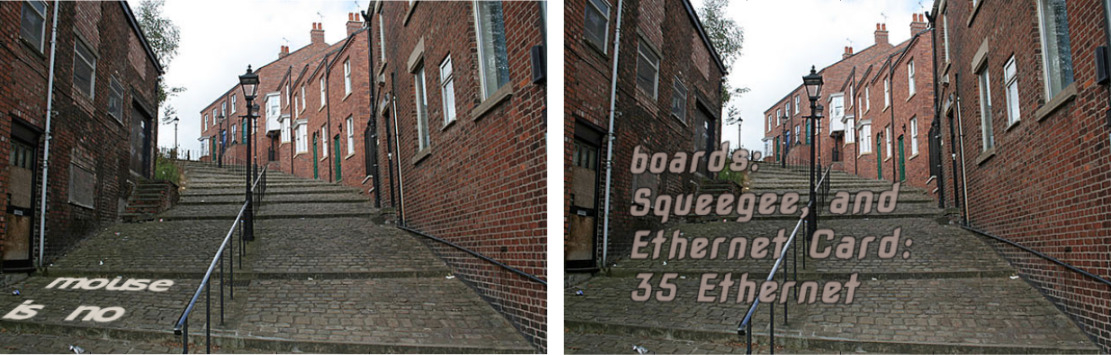}
\caption{Local colour/texture sensitive placement. 
(Left)~Example image from the Synthetic text dataset.  Notice that the
text is restricted within the boundaries of the step in the
street. (Right)~For comparison, the placement of text in this image
does not respect the local region cues.}
\label{fig:seg}
\end{figure}

In real images, text tends to be contained in well defined regions (e.g.\ a sign). We approximate this constraint by requiring text to be contained in regions characterised by a uniform colour and texture. This also prevents text from crossing strong image discontinuities, which is unlikely to occur in practice. Regions are obtained by thresholding the gPb-UCM contour hierarchies~\cite{Arbelaez11} at $0.11$ using the efficient graph-cut implementation of~\cite{Arbelaez14}. Figure~\ref{fig:seg} shows an example of text respecting local region cues.

In natural images, text tends to be painted on top of surfaces (e.g.\ a
sign or a cup). In order to approximate a similar effect in our
synthetic data, the text is perspectively transformed according to
local surface normals. The normals are estimated automatically by
first predicting a dense depth map using the CNN of~\cite{Liu15}
for the regions segmented above, and then fitting a planar facet
to it using RANSAC~\cite{Fischler81}.

Text is aligned to the estimated region orientations as follows: first,  the image region contour is warped to a frontal-parallel view using the estimated plane normal; then, a rectangle is fitted to the fronto-parallel region; finally, the text is aligned to the larger side (``width'') of this rectangle. When placing multiple instances of text in the same region, text masks are checked for collision against each other to avoid placing them on top of each other.

Not all segmentation regions are suitable for text placement --- regions should not be too small, have an extreme aspect ratio, or have surface normal orthogonal to the viewing direction; all such regions are filtered in this stage. Further, regions with too much texture are also filtered, where the degree of texture is measured by the strength of third derivatives in the RGB image.

\paragraph{Discussion.} 
An alternative to using a CNN to estimate depth, which is an error
prone process, is to use a dataset of RGBD images. We prefer to
estimate an imperfect depth map instead because: (1) it allows 
essentially any scene type background image
to be used,
instead of only the ones for
which RGBD data are available, and (2) because publicly available
RGBD datasets such as NYUDv2~\cite{Silberman12}, B3DO~\cite{Janoch11},
Sintel~\cite{Butler12}, and Make3D~\cite{Saxena09} have several
limitations in our context: small size (1,500 images in NYUDv21, 400
frames in Make3D, and a small number of videos in B3DO and Sintel),
low-resolution and motion blur, restriction to indoor images (in
NYUDv2 and B3DO), and limited variability in the images for
video-based datasets (B3DO and Sintel).

\subsection{Text Rendering and Image Composition}\label{sec:color}

Once the location and orientation of text has been decided, text is assigned a colour. The colour palette for text is learned from cropped word images in the IIIT5K word dataset \cite{Mishra12}. Pixels in each cropped word images are partitioned into two sets using K-means, resulting in a colour pair, with one colour approximating the foreground (text) colour and the other the background. When rendering new text, the colour pair whose background colour matches the target image region the best (using L2-norm in the Lab colour space) is selected, and the corresponding foreground colour is used to render the text.

About 20\% of the text instances are randomly chosen to have a border. The border colour is chosen to be either the same as foreground colour with its value channel increased or decreased, or is chosen to be the mean of the foreground and background colours.

To maintain the illumination gradient in the synthetic text image, we blend the text on to the base image using Poisson image editing \cite{Perez03}, with the guidance field defined as in their equation (12). We solve this efficiently using the implementation provided by Raskar\footnote{Fast Poisson image editing code available at: http://web.media.mit.edu/\texttildelow raskar/photo/code.pdf based on Discrete Sine Transform.} 

\section{A Fast Text Detection Network}\label{sec:bbReg}

In this section we introduce our CNN architecture for text detection in natural scenes. While existing text detection pipelines combine several ad-hoc steps and are slow, we propose a detector which is highly accurate, fast, and trainable end-to-end.

Let $\bx$ denote an image.  The most common approach for CNN-based detection is to propose a number of image regions $R$ that may contain the target object (text in our case), crop the image, and use a CNN $c = \phi(\operatorname{crop}_R(\bx)) \in\{0,1\}$ to score them as correct or not. This approach, which has been popularised by R-CNN~\cite{Girshick14}, works well but is slow as it entails evaluating the CNN thousands of times per image. 

An alternative and much faster strategy for object detection is to construct a fixed field of predictors $(c, \bp) = \phi_{uv}(\bx)$, each of which specialises in predicting the presence $c \in \mathbb{R}$ and pose $\bp = (x-u,y-v,w,h)$ of an object around a specific image location $(u,v)$. Here the pose parameters $(x,y)$ and $(w,h)$ denote respectively the location and size of a bounding box tightly enclosing the object. Each predictor $\phi_{uv}$ is tasked with predicting objects which occurs in some ball $(x,y) \in B_\rho(u,v)$ of the predictor location.

While this construction may sound abstract, it is actually a common one, implemented for example by Implicit Shape Models (ISM)~\cite{Leibe04} and Hough voting~\cite{Hough62}. There a predictor $\phi_{uv}$ looks at a local image patch, centred at $(u,v)$,  and tries to predict whether there is an object around $(u,v)$, and where the object is located relative to it.

In this paper we propose an extreme variant of Hough voting, inspired by Fully-Convolutional Network (FCN) of Long~\etal~\cite{Long15} and the You Look Only Once (YOLO) technique of Redmon~\etal~\cite{Redmon16}. In ISM and Hough voting, individual predictions are aggregated across the image, in a voting scheme. YOLO is similar, but avoids voting and uses individual predictions directly; since this idea can accelerate detection, we adopt it here.

The other key conceptual difference between YOLO and Hough voting is that in Hough voting predictors $\phi_{uv}(\bx)$ are local and translation invariant, whereas in YOLO they are not: First, in YOLO each predictor is allowed to pool evidence from the whole image, not just an image patch centred at $(u,v)$. Second, in YOLO predictors at different locations $(u,v)\not=(u',v')$ are different functions $\phi_{uv} \not= \phi_{u'v'}$ learned independently.

While YOLO's approach allows the method to pick up contextual information useful in detection of PASCAL or ImageNet objects, we found this unsuitable for smaller and more variable text occurrences. Instead, we propose here a method which is in between YOLO and Hough voting. As in YOLO, each detector $\phi_{uv}(\bx)$ still predicts directly object occurrences, without undergoing an expensive voting accumulation process; however, as in Hough voting, detectors $\phi_{uv}(\bx)$ are local and translation invariant, sharing parameters. We implement this field of translation-invariant and local predictors as the output of the last layer of a deep CNN, obtaining a \emph{fully-convolutional regression network} (FCRN). 

\setlength{\tabcolsep}{3pt}
\begin{table*}[t]
  \centering
  \resizebox{\textwidth}{!}{%
  \begin{tabular}{|l||cccc|cccc|cccc||ccc|ccc|ccc|}
    \hline
    
    								  & \multicolumn{12}{c||}{\gd PASCAL Eval}  & \multicolumn{9}{c|}{\gd DetEval} \\
								 \cline{2-22} 
	                                           	           & \multicolumn{4}{c|}{\gl IC11}    & \multicolumn{4}{c|}{\gl IC13}    & \multicolumn{4}{c||}{\gl SVT} & \multicolumn{3}{c|}{\gl IC11}    & \multicolumn{3}{c|}{\gl IC13}    & \multicolumn{3}{c|}{\gl SVT}\\   
		           		          		 \cline{2-22}
               							 & F & P &R & $\text{R}_\text{M}$            & F & P &  R &$\text{R}_\text{M}$    &   F & P &R&$\text{R}_\text{M}$ &  F & P &R            & F & P &  R      &F & P &R  \\
    \hline\hline
    Huang~\cite{Huang14}                  &- &- &-&-  &- &- &-&-    &- &- &-&-  & 78 & 88 & 71       &-  & - & -    &- &- &-\\ 
    \hline
    Jaderberg~\cite{Jaderberg15c}                & 77.2 & 87.5 & 69.2&70.6        & 76.2  & 86.7 & 68.0 &69.3          &53.6  &62.8  &46.8 &55.4     & 76.8 & 88.2 & 68.0                   &76.8      & 88.5     & 67.8               &24.7      &27.7      & 22.3\\
    \hline
     \multicell{Jaderberg\\(trained on SynthText)}                   &77.3 &89.2 &68.4 &72.3        & 76.7  & 88.9 & 67.5 & 71.4         &53.6  &58.9 &49.1 &56.1     &75.5 &87.5 &66.4                   &75.5     & 87.9     & 66.3               &24.7    &27.8      &22.3\\
\hline

    Neumann~\cite{Neumann12}                  &- &- &-&-  &- &- &-&-    &- &- &-&-  & 68.7 & 73.1 & 64.7       &-  & - & -    &- &- &-\\ 
    \hline
    Neumann~\cite{Neumann13}                  &- &- &-&-  &- &- &-&-    &- &- &-&-  &72.3 & 79.3 & 66.4        &- &- &-      &- &- &-\\ 
    \hline
     Zhang~\cite{Zhang15}                 &- &- &-&-  &- &- &-&-    &- &- &-&-  & 80 & 84 & {\bf 76}       &80  & 88 & 74    &- &- &-\\
    \hline
    \hline
    FCRN single-scale    					& 60.6 & 78.8 & 49.2 &49.2         & 61.0  & 77.7 & 48.9&48.9          &   45.6& 50.9&41.2& 41.2 & 64.5 &81.9 & 53.2        &  64.3 & 81.3 & 53.1                                & 31.4& {\bf 34.5}& 28.9\\
    
    \hline

    FCRN multi-scale             				& 70.0 & 78.4 & 63.2&64.6         & 69.5  & 78.1 & 62.6&67.0    &46.2& 47.0&45.4&53.0      & 73.0 & 77.9 & 68.9        & 73.4 & 80.3  & 67.7   &   {\bf34.5}& 29.9&{\bf 40.7}\\
    \hline

     FCRN + multi-filt             & 78.7 & {\bf 95.3} & 67.0 & 67.5       & 78.0  & {\bf 94.8} & 66.3&  66.7  &  56.3 & 61.5 & 51.9&54.1 		& 78.0 & {\bf 94.5} & 66.4        &  78.0 & {\bf 94.8} & 66.3   &   25.5& 26.8 & 24.3\\                       
    \hline
     FCRNall + multi-filt             & {\bf 84.7} & 94.3 & {\bf 76.9}& {\bf 79.6}        & {\bf 84.2}  & 93.8 & {\bf 76.4} & {\bf 79.6}   &   {\bf 62.4}& {\bf 65.1} & {\bf 59.9}&{\bf 75.0}           & {\bf 82.3} & 91.5 & 74.8        &  {\bf83.0} & 92.0 & {\bf 75.5}   &   26.7& 26.2 & 27.4\\
    \hline
  \end{tabular}}
\vspace{0.2em}
\caption{\small Comparison with previous methods on text localisation.
		       Precision (P) and Recall (R) at maximum F-measure (F) and the maximum recall ($\text{R}_\text{M}$) are reported.}		  
  \label{table:detect}
\end{table*}

\setlength{\tabcolsep}{1pt}

\subsection{Architecture}\label{s:cnn-architecture}
This section describes the structure of the FCRN. First, we describe the first several
layers of the architecture, which compute text-specific image features.
Then, we describe the dense regression network built on top of these features
and finally its application at multiple scales.

\paragraph{\bf Single-scale features.} Our architecture is inspired 
by VGG-16~\cite{Simonyan15}, using several layers of small dense filters;
however, we found that a much smaller model works just as well and more efficiently for text.
The architecture comprises nine convolutional layers, each followed by the
Rectified Linear Unit non-linearity, and, occasionally, by a max-pooling layer.
All linear filters have a stride of $1$ sample, and preserve the resolution of feature
maps through zero padding. Max-pooling is performed over $2{\times}2$ windows
with a stride of 2 samples, therefore halving the feature maps resolution.
\footnote{The sequence of layers is as follows: 64 5$\times$5 convolutional filters
+ ReLU (CR-64-5$\times$5), max pooling (MP),  CR-128-5$\times$5, MP, CR128-3$\times$3,
CR-128-3$\times$3-conv,  MP, CR-256-3$\times$3,  CR-256-3$\times$3, MP,
CR-512-3$\times$3, CR-512-3$\times$3, CR-512-5$\times$5.}

\paragraph{Class and bounding box prediction.} The single-scale features terminate with
a dense feature field. Given that there are four downsampling max-pooling layers, the stride
of these features is $\Delta = 16$ pixels, each containing 512 feature channels
$\phi_{uv}^f(\bx)$ (we express $uv$ in pixels for convenience).

Given the features $\phi_{uv}^f(\bx)$, we can now discuss the construction of the dense
text predictors $\phi_{uv}(\bx) = \phi_{uv}^r \circ \phi^f(\bx)$. These predictors are implemented
as a further seven $5 \times 5$ linear filters (C-7-5$\times$5) $\phi_{uv}^r$, each regressing
one of seven numbers: the object presence confidence $c$, and up to six object pose
parameters $\bp = (x-u, y-v, w, h, \cos \theta, \sin \theta)$ where $x,y,w,h$ have been
discussed before and $\theta$ is the bounding box rotation.

Hence, for an input image of size $H{\times}W$, we obtain a grid of
$\frac{H}{\Delta}{\times}\frac{W}{\Delta}$ predictions, one each for an image cell of size
$\Delta{\times}\Delta$ pixels.  Each predictor is responsible for detecting a word if the word
centre falls within the corresponding cell.\footnote{For regression, it was found beneficial
to normalise the pose parameters as follows:
$\bar \bp = ((x-u)/\Delta, (y-v)/\Delta, w/W, h/H, \cos \theta, \sin \theta)$.}
 YOLO is similar but operates at about half this resolution; a denser predictor sampling is
 important to reduce collisions (multiple words falling in the same cell) and therefore to
 increase recall (since at most one word can be detected per cell). In practice, for a
 224$\times$224 image, we obtain 14$\times$14 cells/predictors

\paragraph{\bf Multi-scale detection.} Limited receptive field of our convolutional filters
prohibits detection of large text instances.
Hence, we get the detections at multiple down-scaled versions of the input image and
merge them through non-maximal suppression.
In more detail, the input image is scaled down by factors $\{1, 1/2, 1/4, 1/8\}$
(scaling up is an overkill as the baseline features are already computed very densely). 
Then, the resulting detections are combined by suppressing those with a lower score
than the score of an overlapping detection.

\paragraph{Training loss.} We use a squared loss term for each of the
$\frac{H}{\Delta}{\times}\frac{W}{\Delta}{\times}7$ outputs of the CNN as in
YOLO \cite{Redmon16}. If a cell does not contain a ground-truth word, the loss
ignores all parameters but $c$ (text/no-text).
 
\paragraph{Comparison with YOLO.} 
Our fully-convolutional regression network (FCRN) has 30$\times$ less parameters
than the YOLO network (which has ${\sim}90\%$ of the parameters in the last two
fully-connected layers). Due to its global nature, standard YOLO must be retrained for
each image size, including multiple scales, further increasing the model size (while our
model requires 44MB, YOLO would require 2GB). This makes YOLO not only harder to
train, but also less efficient (2$\times$ slower that FCRN).

\section{Evaluation}\label{sec:eval}
First, in \cref{sec:datasets} we describe the text datasets on which we evaluate our model.
Next, we evaluate our model on the text localisation task in \cref{sec:experiments}.
In \cref{sec:ablation}, to investigate which components of the synthetic data generation
pipeline are important, we perform detailed ablation experiments.
In \cref{sec:e2e}, we use the results from our localisation model
for end-to-end text spotting. We show substantial improvements over the
state-of-the-art in both text localisation and end-to-end text spotting.
Finally, in \cref{sec:time} we discuss the speed-up gained by using our models
for text localisation.

\subsection{Datasets}\label{sec:datasets} 
We evaluate our text detection networks on standard
benchmarks: \emph{ICDAR} 2011, 2013
datasets~\cite{Shahab11, Karatzas13} and the Street View Text 
dataset~\cite{Wang10b}. 
These datasets are reviewed next and their
statistics are given in Table~\ref{table:TextDset}.

\paragraph{SynthText in the Wild.} This is a dataset of 800,000
training images generated using our synthetic engine from \cref{sec:synth}.
Each image has about ten word instances annotated with
character and word-level bounding-boxes.

\paragraph{ICDAR Datasets.} The \emph{ICDAR} datasets (IC011, IC013) are obtained from
the Robust Reading Challenges held in 2011 and 2013 respectively. They contain real
world images of text on sign boards, books, posters and other objects with world-level
axis-aligned bounding box annotations. The datasets largely contain the same images,
but shuffle the test and training splits. We do not evaluate on the more recent \emph{ICDAR}
2015 dataset as it is almost identical to the 2013 dataset.

\paragraph{Street View Text.}  This dataset, abbreviated \emph{SVT}, consists of images
harvested from Google Street View annotated with word-level axis-aligned bounding boxes.
\emph{SVT} is more challenging than the \emph{ICDAR} data as it contains
smaller and lower resolution text.
Furthermore, not all instances of text are annotated. In practice, this means that
precision is heavily underestimated in evaluation.  Lexicons consisting of 50 distractor
words along with the ground-truth words are provided for each image; we refer to
testing on \emph{SVT} with these lexicons as \emph{SVT-50}.

\subsection{Text Localisation Experiments}\label{sec:experiments}
We evaluate our detection networks to~---~
(1)~compare the performance when applied to single-scale and multiple down-scaled
versions of the image and,
(2)~improve upon the state-of-the-art results in text detection
when used as high-quality proposals.

\paragraph{Training.}
FCRN is trained on 800,000 images from our \emph{SynthText~in~the~Wild} dataset.
Each image is resized to a size of 512$\times$512 pixels. We optimise using SGD
with momentum and batch-normalisation \cite{Ioffe15} after every convolutional 
layer (except the last one). We use mini-batches of 16 images each, set the 
momentum to 0.9, and use a weight-decay of $5^{-4}$.
The learning rate is set to $10^{-4}$ initially and is reduced to $10^{-5}$
when the training loss plateaus.

As only a small number (1-2\%) of grid-cells contain text, we weigh
down the non-text probability error terms initially by multiplying with 0.01;
this weight is gradually increased to 1 as the training progresses.
Due to class imbalance, all the probability scores collapse to zero
if such a weighting scheme is not used.

\paragraph{Inference.}
%
We get the class probabilities and
bounding-box predictions from our FCRN model. The predictions are
filtered by thresholding the class probabilities (at a threshold $t$).  
Finally,
multiple detections from nearby cells are suppressed using non-maximal
suppression, whereby amongst two overlapping detections the one with
the lower probability is suppressed.
In the following we first give results for a conservative threshold of 
$t = 0.3$, for higher precision, and then relax this to $t = 0.0$ (i.e.,\
all proposals accepted) for higher recall.

\paragraph{Evaluation protocol.}
%
We report text detection 
performance using two protocols commonly used in the literature ---~
(1)~\emph{DetEval}~\cite{Wolf06} popularly used in ICDAR competitions for
evaluating localisation methods, and (2)~PASCAL VOC style
intersection-over-union overlap method ($\geq$ 0.5 IoU for a positive detection).

\paragraph{Single \& multi-scale detection.}
%
The ``FCRN single-scale'' entry in Table~\ref{table:detect} shows the
performance of our FCRN model on the test datasets.
The precision at maximum F-measure of single-scale FCRN
is comparable to the methods of Neuman~\etal~\cite{Neumann12, Neumann13},
while the recall is significantly worse by 12\%.

The ``FCRN multi-scale'' entry in Table~\ref{table:detect} shows
performance on multi-scale application of our network.
This method improves maximum recall by more than 12\%
over the single-scale method and outperforms the methods of Neumann~\etal.

\begin{figure}[t]
\centering
\includegraphics[width=0.8\linewidth]{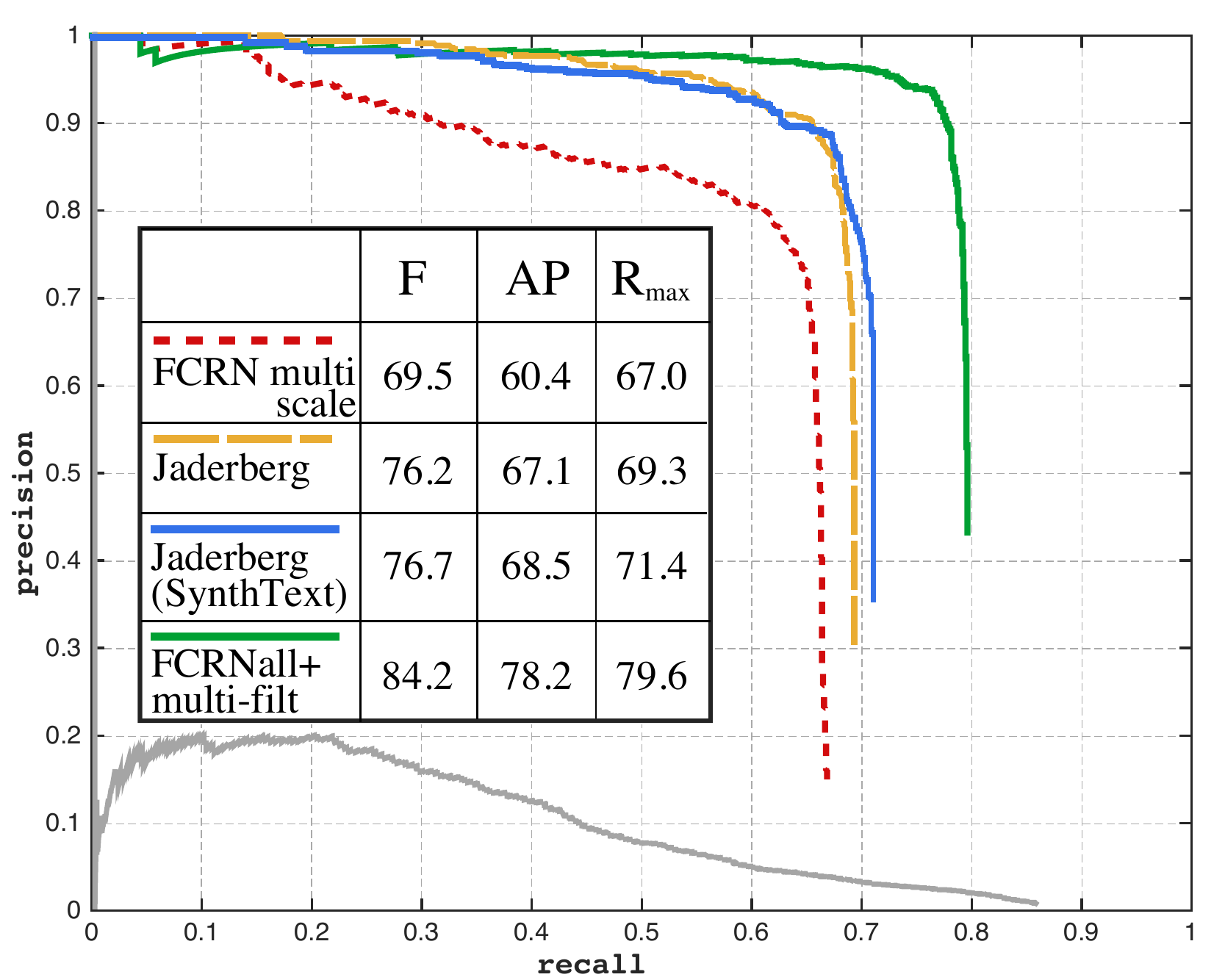}
\caption{
\small Precision-Recall curves 
for various text detection methods on IC13. 
The methods are:
(1)~multi-scale application of FCRN
(``FCRN-multi''); (2) The original curve of Jaderberg~\etal~\cite{Jaderberg15c};
(3) Jaderberg~\etal~\cite{Jaderberg15c} retrained on 
the \emph{SynthText in the Wild} dataset;  and, (4)
``FCRNall + multi-filt'' methods. 
Maximum
F-score (F), Average Precision (AP) and maximum Recall
($\text{R}_\text{max}$) are also given. The gray curve at the
bottom is of multi-scale detections from our FCRN network (max. recall
= 85.9\%), which is fed into the multi-filtering post-processing to get
the refined ``FCRNall + multi-filt'' detections.}
\label{fig:pr}
\end{figure}

\paragraph{Post-processing proposals.}
Current end-to-end text spotting (detection and recognition) methods
\cite{Alsharif14, Jaderberg15c, Wang12} boost performance by combining detection
with text recognition. To further improve FCRN detections, we 
use the multi-scale detections from FCRN as proposals and refine
them by using the post-processing stages of Jaderberg~\etal~\cite{Jaderberg15c}.
There are three stages: first filtering using a
binary text/no-text random-forest classifier; second, regressing
an improved bounding-box using a CNN; and third recognition based NMS
where the word images are recognised using a large fixed
lexicon based CNN, and the detections are merged through
non-maximal suppression based on word identities. Details are
given in~\cite{Jaderberg15c}. We use code provided by the authors for fair
comparison.

We test this in two modes ~---~(1) ~\emph{low-recall}: where only
high-scoring (probability $>0.3$) multi-scale FCRN detections are used
(the threshold previously used in the single- and multi-scale
inference). This typically yields less than 30 proposals.  And,
(2)~\emph{high-recall}: where {\em all} the multi-scale FCRN detections
(typically about a thousand in number) are used.  Performance of these
methods on text detection are shown by the entries named ``FCRN +
multi-filt'' and ``FCRNall + multi-filt'' respectively in
Table~\ref{table:detect}.  Note that the
\emph{low-recall} method achieves better than the state-of-the-art
performance on text detection, whereas \emph{high-recall} method
significantly improves the state-of-the-art with an improvement of 6\%
in the F-measure for all the datasets.

\begin{figure}[t]
\centering
\includegraphics[width=0.8\linewidth]{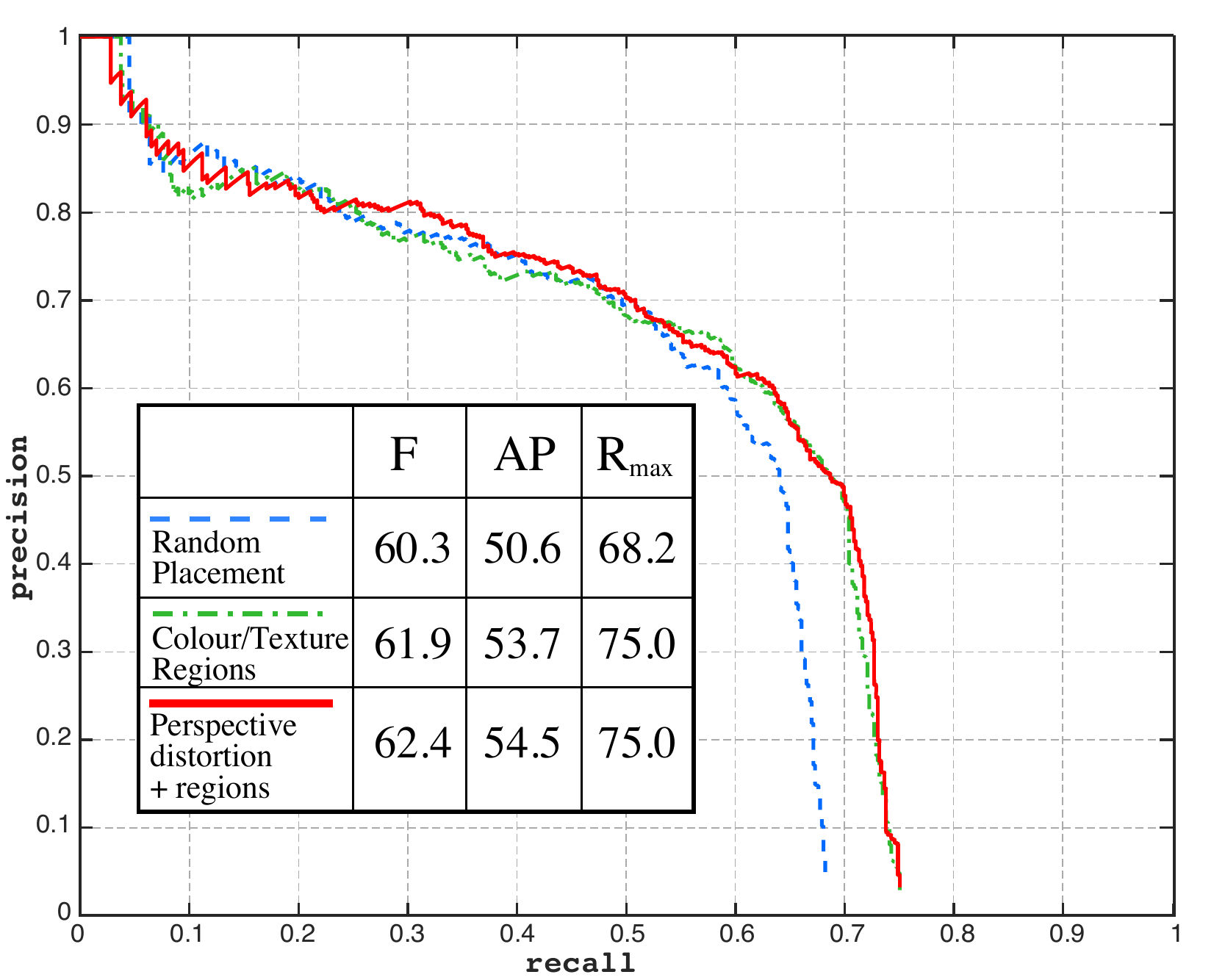}
\caption{\small Precision-Recall curves text localisation on the SVT dataset
using the model ``FCRNall+multi-filt'' when trained on increasingly sophisticated training sets (\cref{sec:ablation}).}
\label{fig:ablation}
\end{figure}

Figure~\ref{fig:pr} shows the Precision-Recall curves for text
detection on the \emph{IC13} dataset. 
Note the high recall (85.9\%) of 
the multi-scale
detections output from FCRN before refinement using the
multi-filtering post-processing.
Also, note 
the drastic increase in maximum recall ($+10.3\%$) and in Average
Precision ($+11.1\%$) for ``FCRNall + multi-filt'' as compared to Jaderberg~\etal.

 Further, to establish that the improvement in text detection
is due to the new detection model, and not merely due to the large
size of our synthetic dataset, we trained Jaderberg~\etal's method
on our \emph{SynthText in the Wild} dataset -- in particular, the ACF
component of their region proposal stage.\footnote{Their other region proposal
method, EdgeBoxes, was not re-trained; as it is learnt from low-level edge
features from the Berkeley Segmentation Dataset, which is not text specific.}
Figure~\ref{fig:pr} and Table~\ref{table:detect} show that, even with 10$\times$ more (synthetic)
training data, Jaderberg~\etal's model improves only marginally
(+0.8\% in AP, +2.1\% in maximum recall).

A common failure mode is text in unusual fonts which are not present in the training set.
The detector is also confused by symbols or patterns of constant stroke width which look
like text, for example road-signs, stick figures etc. Since the detector does not scale
the image up, extremely small sized text instances are not detected. Finally, words 
get broken into multiple instances or merged into one instance due to large or small spacing between the characters.

 \subsection{Synthetic Dataset Evaluation}\label{sec:ablation} 
 
 We investigate the contribution that the various stages of the synthetic text-scene 
 data generation pipeline bring to localisation accuracy: We generate three synthetic
 training datasets with increasing levels of sophistication, where the text is~
 (1) is placed at random positions within the image,
 (2) restricted to the local colour and texture boundaries, and
 (3) distorted perspectively to match the local scene depth
 (while also respecting the local colour and texture boundaries as in (2) above).
 All other aspects of the datasets were kept the same --- e.g.\ the text lexicon,
 background images, colour distribution.
 
Figure~\ref{fig:ablation} shows the results on localisation on the \emph{SVT}
dataset of our method ``FCRNall+multi-filt''.  Compared to random placement,
restricting text to the local colour and texture regions
significantly increases the maximum recall (+6.8\%), AP (+3.85\%), and the maximum
F-measure (+2.1\%). Marginal improvements are seen with the addition
of perspective distortion: +0.75\% in AP, +0.55\% in maximum F-measure, and no change
in the maximum recall. This is likely due to the fact that most text instances in the \emph{SVT}
datasets are in a fronto-parallel orientation. Similar trends are
observed with the \emph{ICDAR} 2013 dataset, but with more contained differences probably due to the fact that \emph{ICDAR}'s text instances
are much simpler than \emph{SVT}'s and benefit less from the more advanced datasets.

\begin{table}[t]
\begin{center}
\begin{tabular}[t]{|l||c|c|c|c|c|}
\hline
Model&                                      IC11    &  IC11* &    IC13 &  SVT & SVT-50\\
\hline\hline
Wang~\cite{Wang11} &                  -      &    -      &      -               &     -    &  38   \\
Wang \& Wu~\cite{Wang12} &       -      &    -      &      -               &     -    &  46   \\
Alsharif~\cite{Alsharif14}  &           -      &    -      &      -               &     -    &  48   \\
Neumann~\cite{Neumann13} &     -      &   45.2    &     -                &     -    &    -    \\
Jaderberg~\cite{Jaderberg14}&     -      &    -      &      -               &     -    &  56\\
Jaderberg~\cite{Jaderberg15c} &        76   &  69     &    76               &   53    &  {\bf 76}\\
\hline
\shortstack{FCRN + multi-filt} &     \multicellC{80.5\\(77.8)} & \multicellC{75.8\\(73.5)} & \multicellC{80.3\\(77.8)}         & 54.7 &  68.0\\
\hline
\shortstack{FCRNall + multi-filt}&  \multicellC{{\bf 84.3}\\(81.2)} & \multicellC{{\bf 81.0}\\(78.4)} & \multicellC{{\bf 84.7}\\(81.8)}         &{\bf 55.7}  &  67.7\\
\hline
\end{tabular}
\end{center}
\caption{\small 
Comparison with previous methods on end-to-end text spotting.  Maximum
F-measure\% is reported.  IC11* is evaluated according to the protocol
described in~\cite{Neumann13}. Numbers in parenthesis are
obtained if words containing non-alphanumeric characters are not ignored -- SVT does not have any of these.}
\vspace{-0.5em}
\label{table:e2e}
\end{table}

\subsection{End-to-End Text Spotting}\label{sec:e2e}
Text spotting is limited by the detection stage, as state-of-the-art cropped word
image recognition accuracy is over 98\%~\cite{Jaderberg14c}. We utilise our
improvements in text localisation to obtain state-of-the-art results in text spotting.

\paragraph{Evaluation protocol.}
%
Unless otherwise stated,
we follow the standard evaluation protocol by Wang~\etal~\cite{Wang11},
where all words that are either less than three characters long
or contain non-alphanumeric characters are ignored.
An overlap (IoU) of at least 0.5 is required for a positive detection.

Table~\ref{table:e2e} shows the results on end-to-end text spotting task using
the ``FCRN + multi-filt'' and  ``FCRNall~+~multi-filt'' methods.
For recognition we use the output of the intermediary recognition stage of the 
pipeline based on the lexicon-encoding CNN of Jaderberg~\etal~\cite{Jaderberg14c}.
We improve upon previously reported results (F-measure):
$+8$\% on the \emph{ICDAR} datasets, and $+3$\% on the \emph{SVT} dataset.
Given the high recall of our method (as noted before in Figure~\ref{fig:pr}),
the fact that many text instances are unlabelled in \emph{SVT} cause precision to drop; 
hence, we see smaller gains in \emph{SVT} and do worse on \emph{SVT-50}.

\subsection{Timings}\label{sec:time}
%
At test time FCRN can process 20 images per second (of size
$512{\times}512$px) at single scale and about 15 images per second
when run on multiple scales (1,1/2,1/4,1/8) on a GPU. When used as
high-quality proposals in the text localisation pipeline of
Jaderberg~\etal~\cite{Jaderberg15c}, it replaces the region proposal stage
which typically takes about 3 seconds per image. Hence, we gain a
speed-up of about 45 times in the region proposal stage. Further, the
``FCRN + multi-filt'' method, which uses only the high-scoring
detections from multi-scale FCRN and achieves state-of-the-art results
in detection and end-to-end text spotting, cuts down the number of
proposals in the later stages of the pipeline by a factor of 10:
the region proposal stage of Jaderberg~\etal proposes about 2000 boxes
which are quickly filtered using a random-forest classifier to a
manageable set of about 200 proposals, whereas the high-scoring
detections from multi-scale FCRN are typically less than 30.
Table~\ref{table:time} compares the time taken for end-to-end
text-spotting; our method is between 3$\times$ to 23$\times$
faster than Jaderberg~\etal's, depending on the variant.

\begin{table}[t]
\begin{center}
\resizebox{\linewidth}{!}{
\begin{tabular}{|l||c||c|c|c|}
	\hline
	&    \gl Total Time  &   \gl \shortstack{Region\\Proposal}  &  \gl \shortstack{Proposal\\Filtering}  &  \gl\shortstack{BB-regression\\\& recognition} \\
	\hline
	FCRN+multi-filt     &  0.30 &  0.07 &  0.03 & 0.20\\
	FCRNall+multi-filt &  2.47 &  0.07 &  1.20 & 1.20\\
	Jaderberg~\etal    &  7.00 &  3.00 &  3.00 & 1.00\\
	\hline
\end{tabular}}
\end{center}
\caption{\small Comparison of end-to-end text-spotting time (in seconds).}
\vspace{-0.5em}
\label{table:time}
\end{table}

\section{Conclusion}\label{sec:conclude}

We have developed a new CNN architecture
for generating text proposals in images. It would not have been possible
to train this architecture on the available annotated datasets, as they contain far
too few samples, but we have shown that training images of sufficient
verisimilitude can be generated synthetically, and that the CNN
trained {\em only} on these images exceeds the state-of-the-art
performance for both detection and end-to-end text spotting on real
images.

\paragraph{Acknowledgements.}
%
We thank Max Jaderberg for generously providing code and helpful advice.
We are grateful for comments from Jiri Matas. Financial support was provided by
the UK EPSRC CDT in Autonomous
Intelligent Machines and Systems Grant EP/L015987/2, EPSRC Programme
Grant Seebibyte EP/M013774/1, and the Clarendon Fund scholarship. 

{\small
\bibliographystyle{ieee}
\bibliography{shortstrings,vgg_local,vgg_other}
}

\clearpage
\onecolumn

\begin{appendices}
\section{Appendix}
We highlight some components of our synthetic text dataset in sections~\ref{sec:comp} and~\ref{sec:blend}, 
and show some sample images from the dataset in~\cref{sec:samples}.
Finally, we compare the detection results from our ``FCRNall multi-filt'' method
and Jaderberg et al.~\cite{Jaderberg15c} on the ICDAR 2013 dataset in~\cref{sec:showICDAR}
and the Street View Text (SVT) dataset in~\cref{sec:showSVT}.

\vfill

\subsection{Variation in Fonts, Colors and Sizes}\label{sec:comp}
The following images show synthetic text renderings for the same text -- ``vamos!''.\\
Along the rows, the text is rendered in approximately the same location
and against the same background image but in different fonts, colours and sizes.
\setlength{\tabcolsep}{1pt}
\begin{table}[h]
\centering
\resizebox*{\textwidth}{!}{%
\begin{tabular}{ccc}
\includegraphics[width=0.32\textwidth]{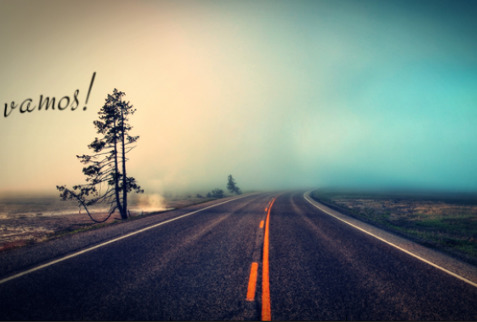} &
\includegraphics[width=0.32\textwidth]{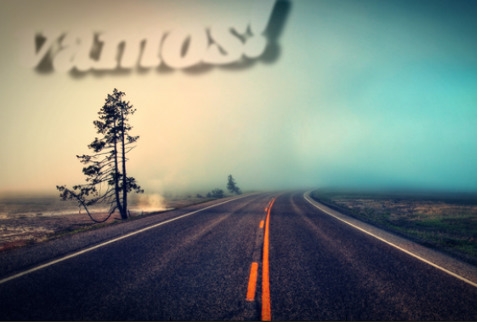} &
\includegraphics[width=0.32\textwidth]{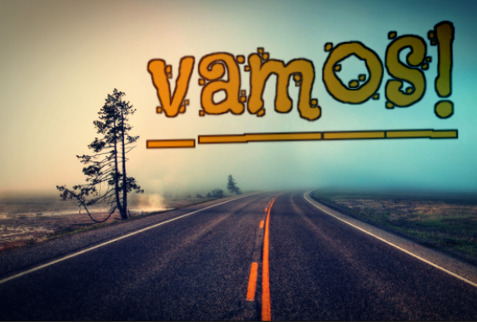}\\
\includegraphics[width=0.32\textwidth]{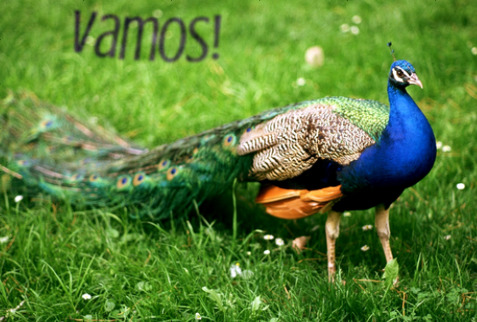} &
\includegraphics[width=0.32\textwidth]{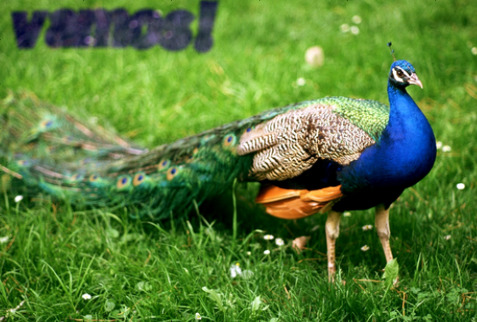} &
\includegraphics[width=0.32\textwidth]{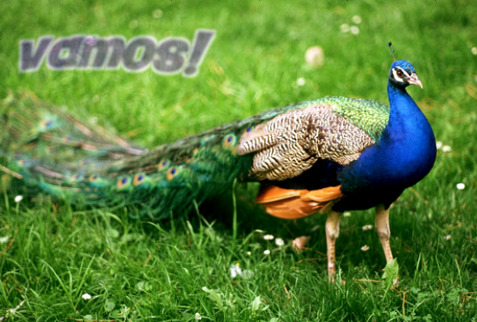}\\
\includegraphics[width=0.32\textwidth]{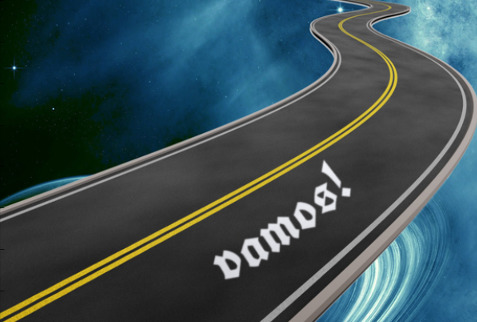} &
\includegraphics[width=0.32\textwidth]{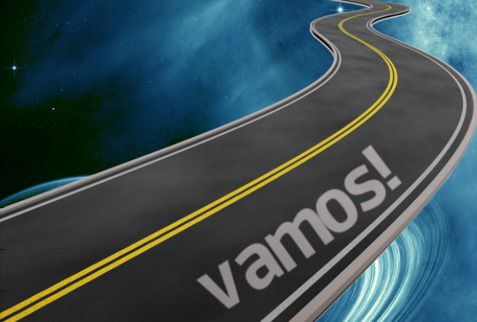} &
\includegraphics[width=0.32\textwidth]{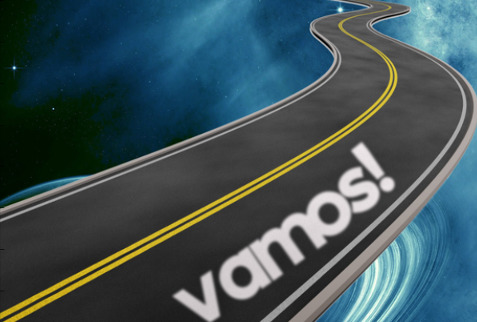}
\end{tabular}}
\end{table}
\vfill
\clearpage

\subsection{Poisson Editing vs. Alpha Blending}\label{sec:blend}
Comparison between simple alpha blending {\bf (bottom row)} and Poisson
Editing~\cite{Perez03} {\bf (top row)}.\\
Poisson Editing preserves local illumination gradient and texture details.

\setlength{\tabcolsep}{1pt}
\begin{table}[h]
\centering
\resizebox{\textwidth}{!}{%
\begin{tabular}{ccccc}
\rotatebox{90}{\hspace{0.6cm}Poisson Editing} &&
\includegraphics[width=0.32\textwidth]{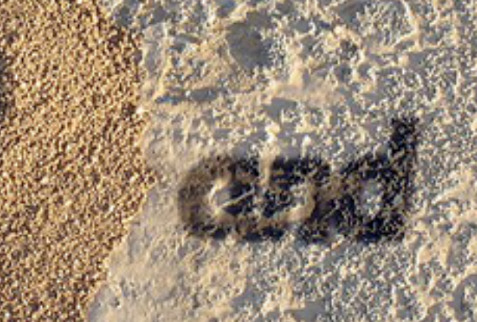} &
\includegraphics[width=0.32\textwidth]{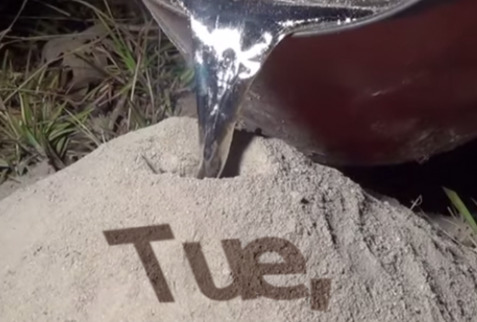} &
\includegraphics[width=0.32\textwidth]{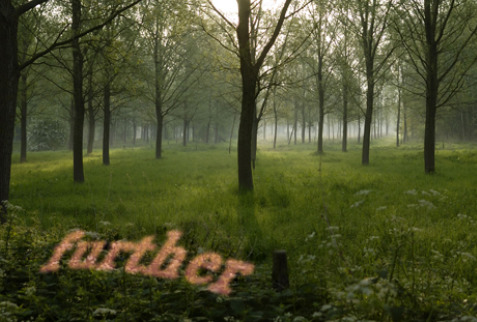}\\\\
\rotatebox{90}{\hspace{0.7cm}Alpha Blending}&&
\includegraphics[width=0.32\textwidth]{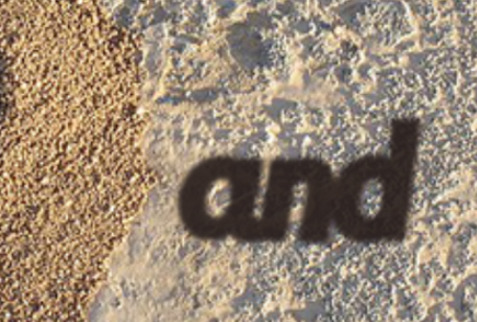} &
\includegraphics[width=0.32\textwidth]{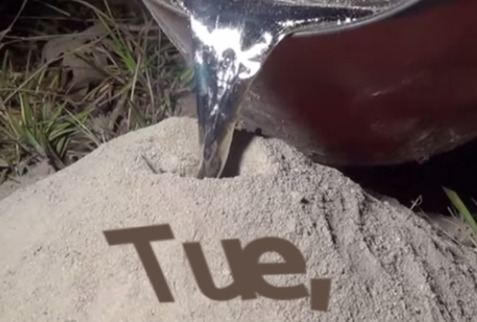} &
\includegraphics[width=0.32\textwidth]{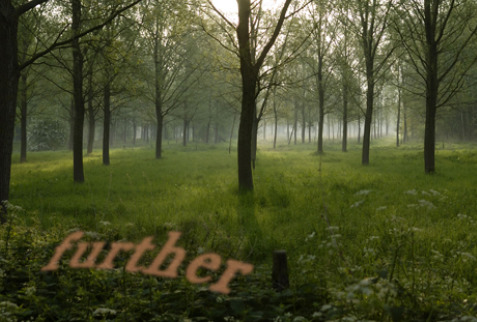}
\end{tabular}}
\end{table}

\subsection{SynthText in the Wild}\label{sec:samples}
Sample images from our synthetic text dataset (continued on the next page).\\
These images show text instances in various fonts, colours, sizes, with borders
and shadows, against different backgrounds, and transformed according to the
local geometry and constrained to local contiguous regions of colour and text.
Ground-truth word bounding-boxes are marked in {\color{red} red}.

\setlength{\tabcolsep}{1pt}
\begin{table}[h]
\centering
\resizebox*{\textwidth}{!}{
\begin{tabular}{ccc}
\includegraphics[width=0.32\textwidth]{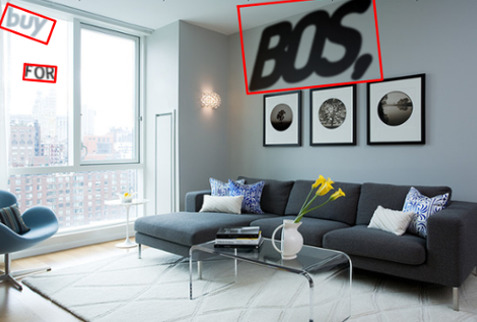} &
\includegraphics[width=0.32\textwidth]{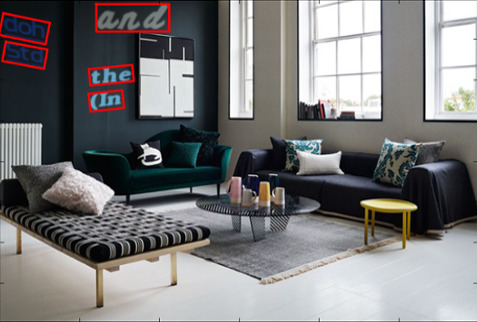} &
\includegraphics[width=0.32\textwidth]{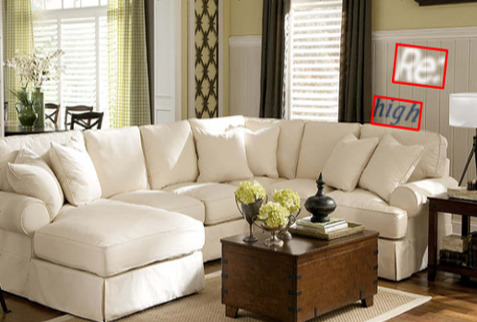} \\
\includegraphics[width=0.32\textwidth]{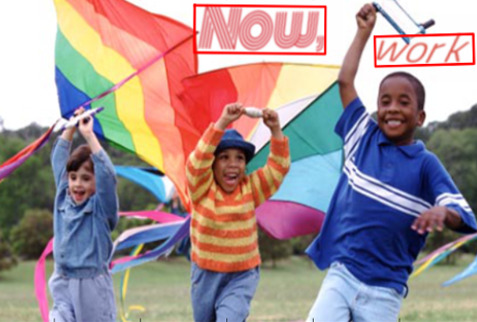} &
\includegraphics[width=0.32\textwidth]{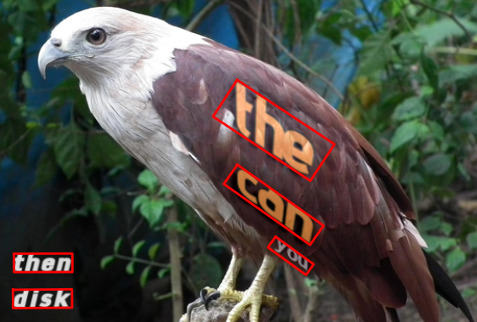} &
\includegraphics[width=0.32\textwidth]{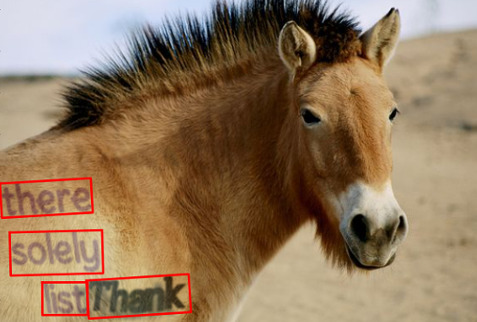}
\end{tabular}}
\end{table}
\clearpage

\begin{table}[h]
\centering
\resizebox*{\textwidth}{\textheight}{%
\begin{tabular}{ccc}
\includegraphics{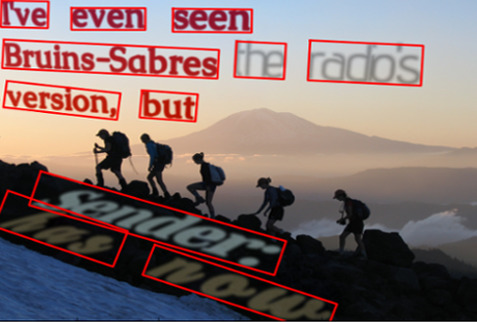} &
\includegraphics{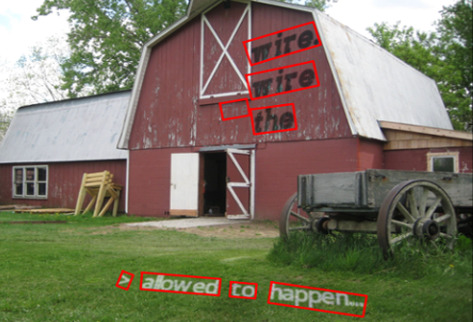} &
\includegraphics{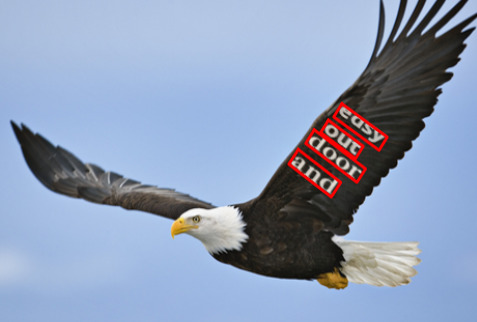} \\
\includegraphics{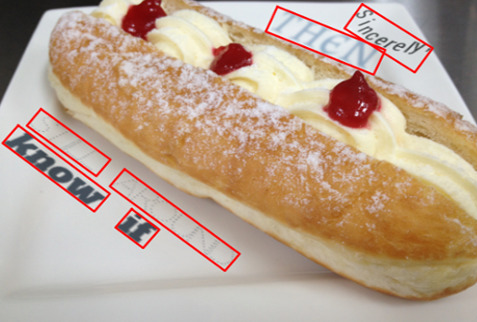} &
\includegraphics{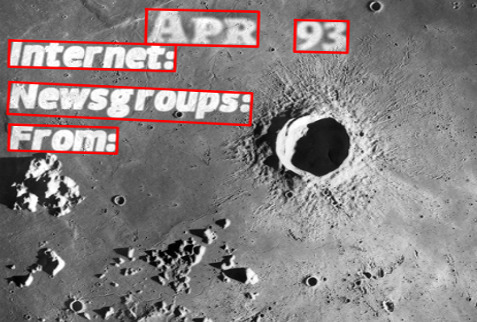} &
\includegraphics{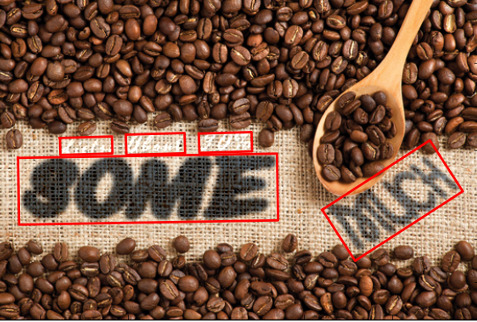} \\
\includegraphics{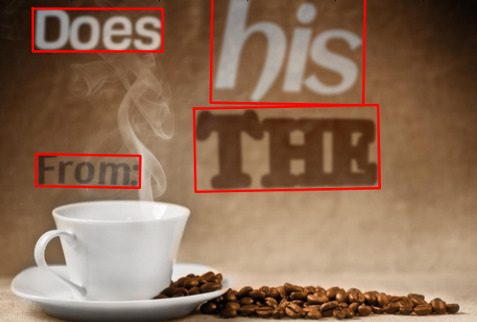} &
\includegraphics{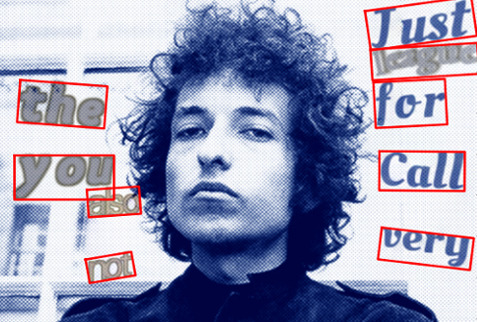} &
\includegraphics{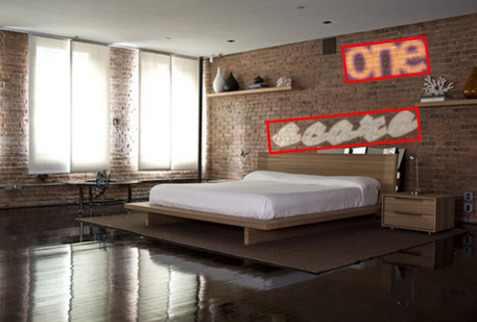} \\
\includegraphics{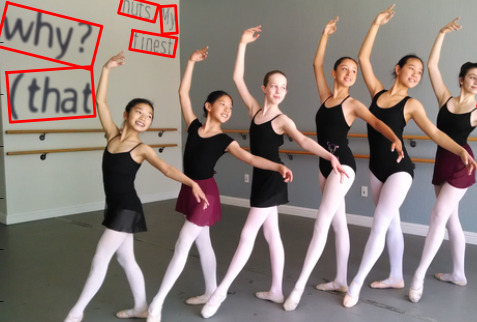} &
\includegraphics{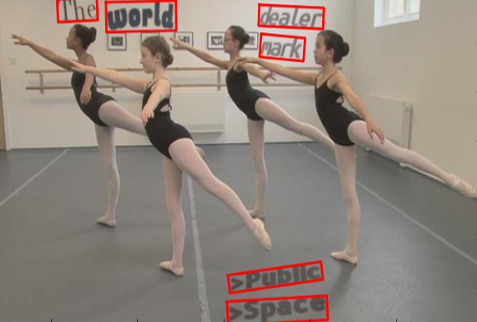} &
\includegraphics{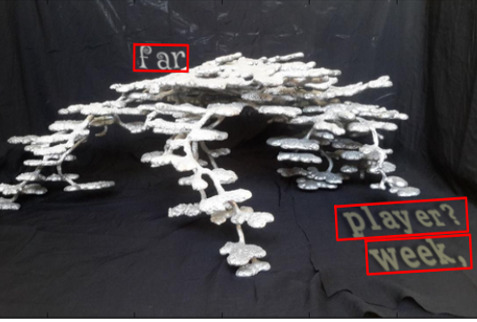} \\
\includegraphics{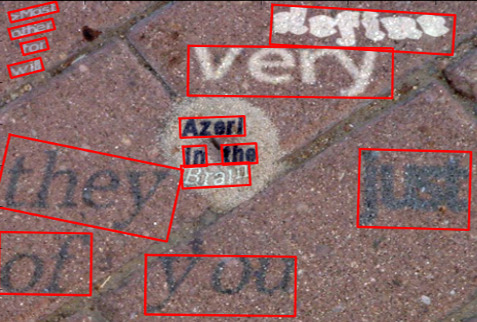} &
\includegraphics{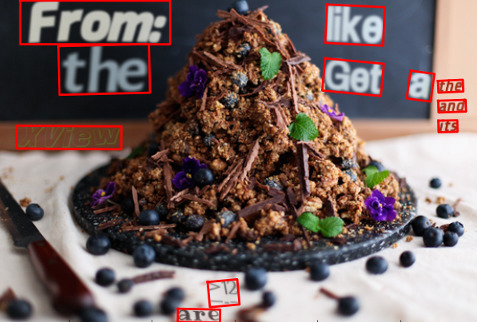} &
\includegraphics{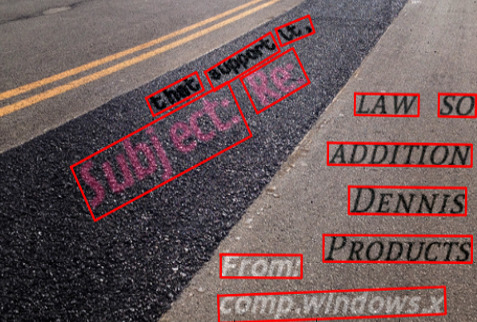} 
\end{tabular}}
\end{table}
\clearpage

\subsection{ICDAR 2013 Detections}\label{sec:showICDAR}
Example detections on the ICDAR 2013 dataset from ``FCRNall + multi-flit''~{\bf (top row)} and
those from Jaderberg~\emph{et al.}~\cite{Jaderberg15c}~{\bf (bottom row)}.
Precision, recall and F-measure values {\bf (P/R/F)} are indicated at the top of each image.
\setlength{\tabcolsep}{1pt}
\begin{table}[h]
\centering
\resizebox*{\textwidth}{!}{%
\begin{tabular}{ccccccc}
&&1 / 1 / 1  & 1 / 1 / 1 & 1 / 0.88 / 0.94 & 1 / 1 / 1 & 1 / 1 / 1\\
\rotatebox{90}{\hspace{0.5cm}FCRNall multi-filt} &&
\includegraphics[width=0.2\textwidth]{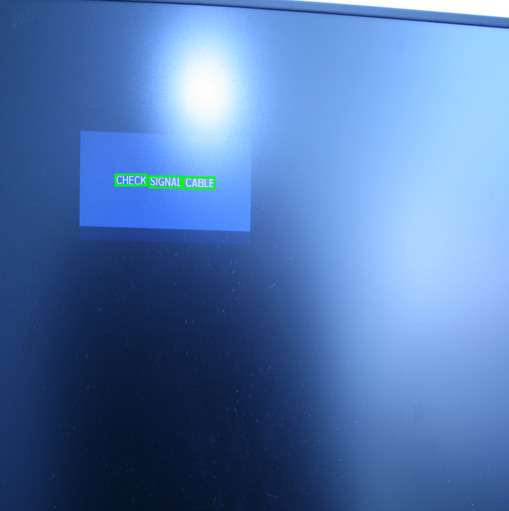} &
\includegraphics[width=0.2\textwidth]{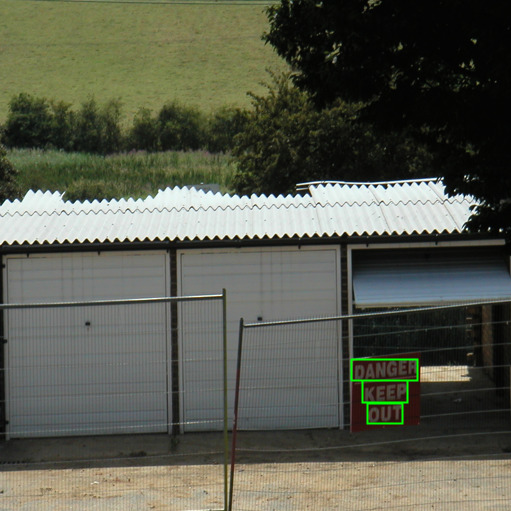} &
\includegraphics[width=0.2\textwidth]{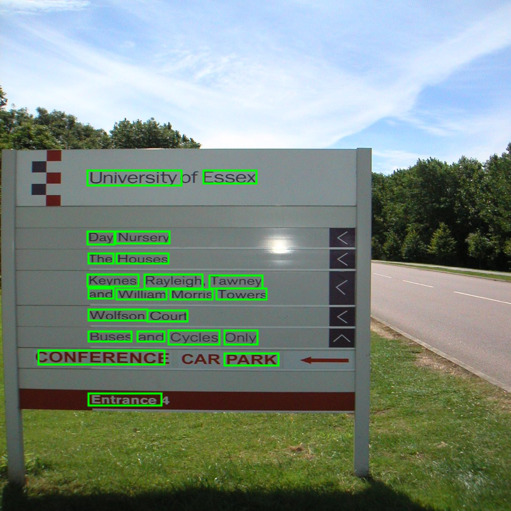} &
\includegraphics[width=0.2\textwidth]{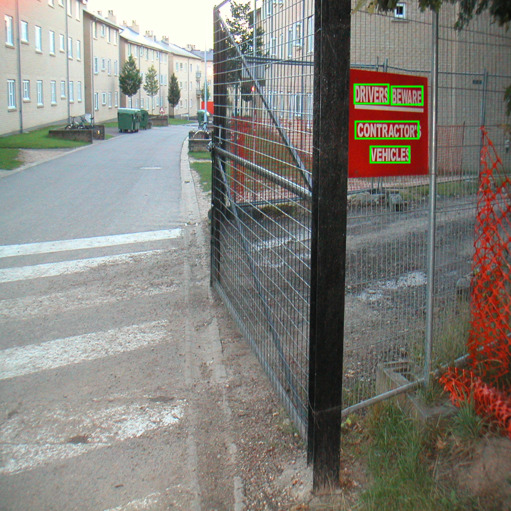} &
\includegraphics[width=0.2\textwidth]{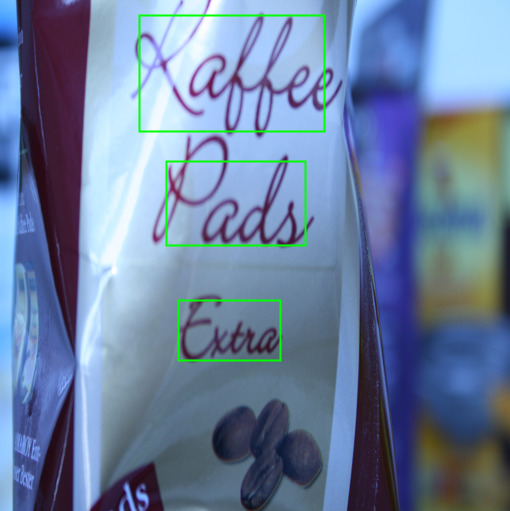} \\\\

&&1 / 0.33 / 0.50 & 1 / 0.33 / 0.50 & 1 / 0.28 / 0.44 & 1 / 0.75 / 0.86 & 1 / 0.66 / 0.80\\
\rotatebox{90}{\hspace{0.7cm}Jaderberg~\emph{et al.}} &&
\includegraphics[width=0.2\textwidth]{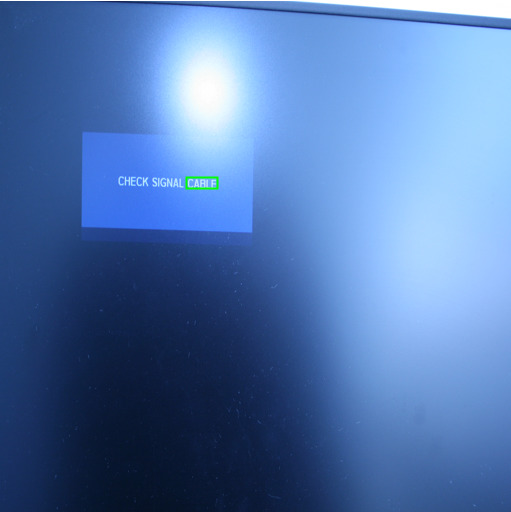} &
\includegraphics[width=0.2\textwidth]{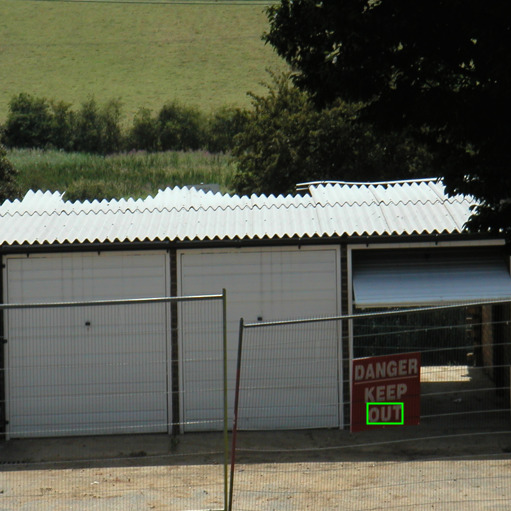} &
\includegraphics[width=0.2\textwidth]{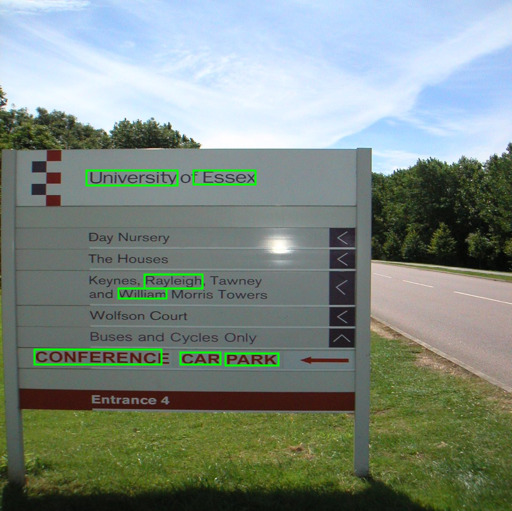} &
\includegraphics[width=0.2\textwidth]{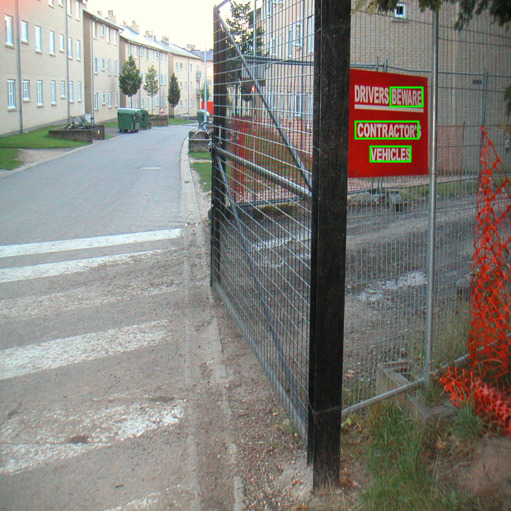} &
\includegraphics[width=0.2\textwidth]{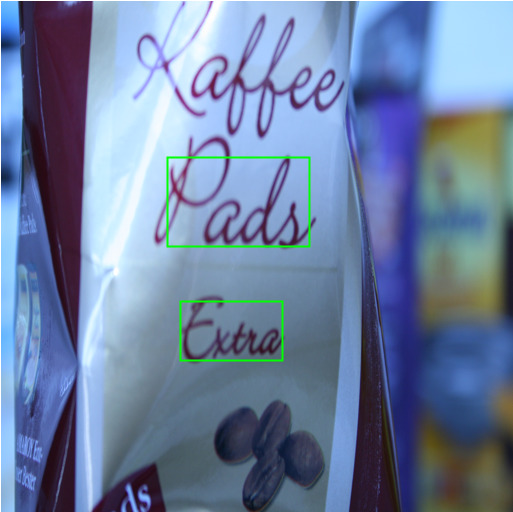} 
\end{tabular}}
\end{table}

\subsection{Street View Text (SVT) Detections}\label{sec:showSVT}
Example detections on the Street View Text (SVT) dataset from ``FCRNall + multi-flit''~{\bf (top row)}
and those from Jaderberg~\emph{et al.}~\cite{Jaderberg15c}~{\bf (bottom row)}.
Precision, recall and F-measure values {\bf (P/R/F)} are indicated at the top of each image:
both the methods have a precision of 1 on these images
(except in one case due to missing ground-truth annotation).

\setlength{\tabcolsep}{1pt}
\begin{table}[h]
\centering
\resizebox*{\textwidth}{!}{%
\begin{tabular}{cccccccc}
&&1 / 1 / 1  & 1 / 1 / 1 & 1 / 1 / 1 & 0.80 / 1 / 0.89 & 1 / 1 / 1\\
\rotatebox{90}{\hspace{0.5cm}FCRNall multi-filt} &&
\includegraphics[width=0.2\textwidth]{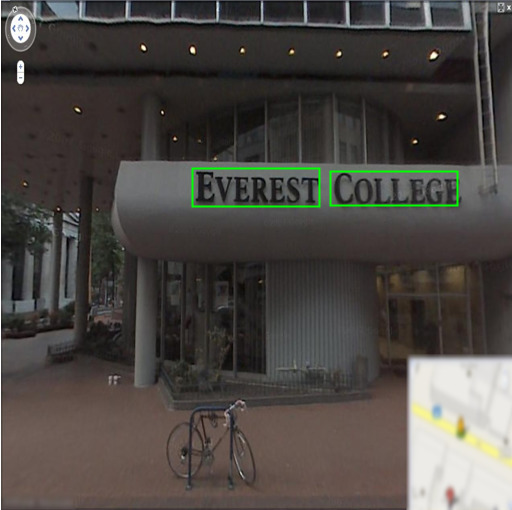} &
\includegraphics[width=0.2\textwidth]{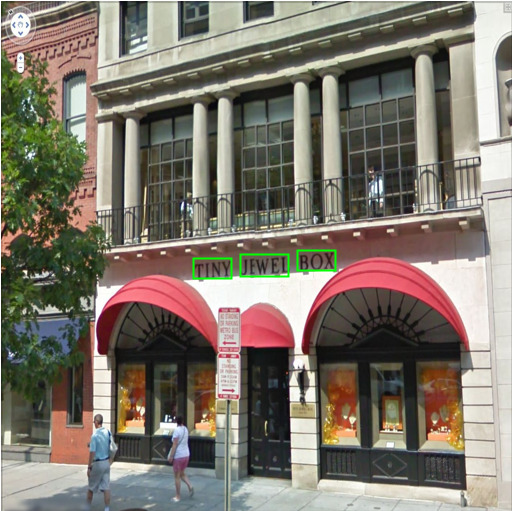} &
\includegraphics[width=0.2\textwidth]{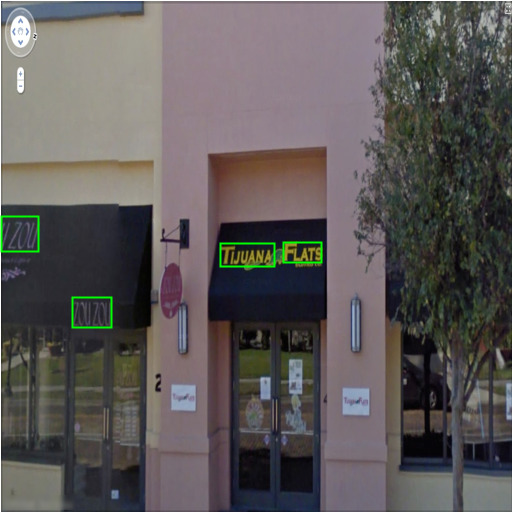} &
\includegraphics[width=0.2\textwidth]{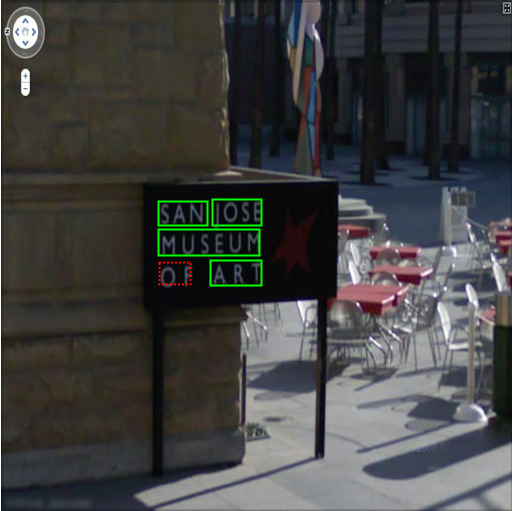} &
\includegraphics[width=0.2\textwidth]{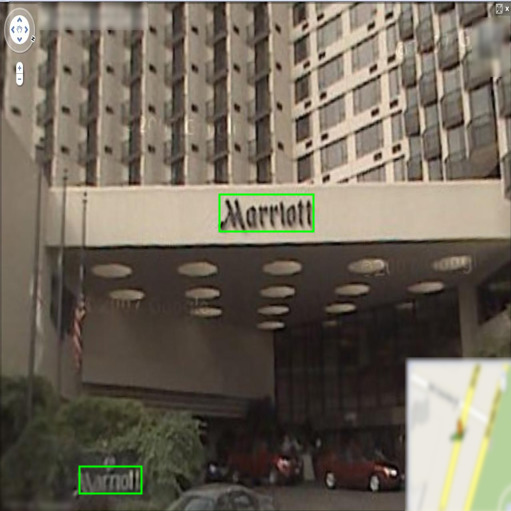} \\\\

&&1 / 1 / 1 & 1 / 0 / 0 & 1 / 0.5 / 0.67 & 0.75 / 0.75 / 0.75 & 1 / 0.5 / 0.67\\
\rotatebox{90}{\hspace{0.7cm}Jaderberg~\emph{et al.}} &&
\includegraphics[width=0.2\textwidth]{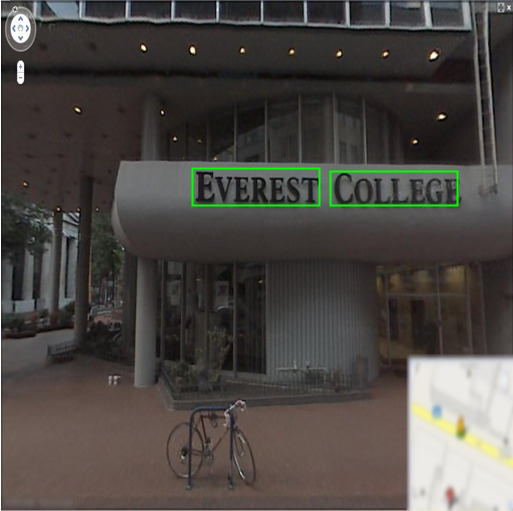} &
\includegraphics[width=0.2\textwidth]{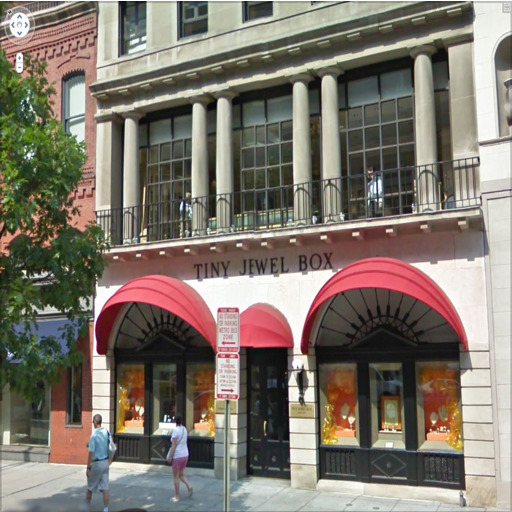} &
\includegraphics[width=0.2\textwidth]{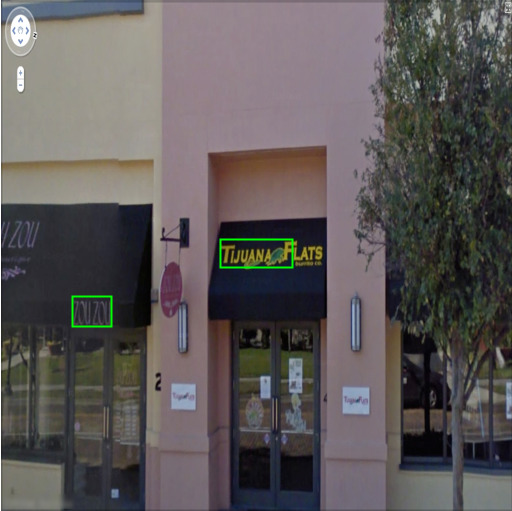} &
\includegraphics[width=0.2\textwidth]{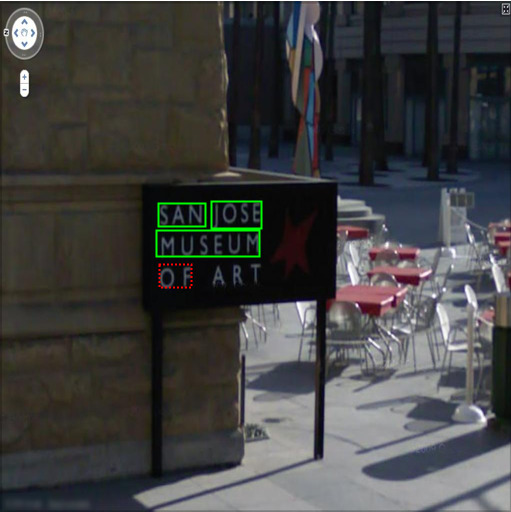} &
\includegraphics[width=0.2\textwidth]{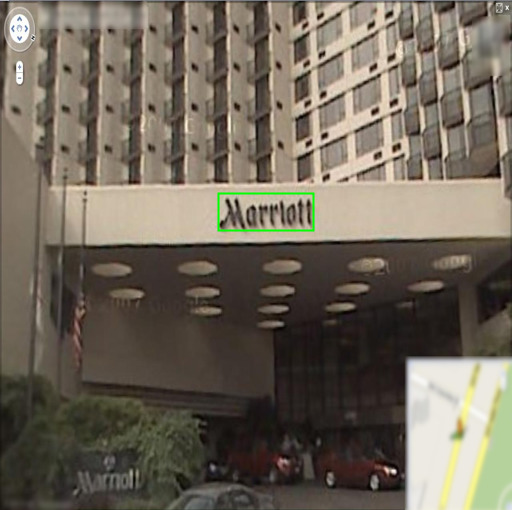} 
\end{tabular}}
\end{table}

\end{appendices}

\end{document}